\documentclass[authoryear,preprint,review,12pt,natbib]{elsarticle}

\usepackage{amssymb}
\usepackage{amsmath}

\usepackage{booktabs}
\usepackage{float}
\usepackage{graphicx}

\usepackage{lineno}

\usepackage[hidelinks]{hyperref} 

\usepackage[commandnameprefix=always,final]{changes} 

\definechangesauthor[name=Ayrat Abdullin, color=green]{AA}
\definechangesauthor[name=Denis Anikiev, color=magenta]{DA}
\definechangesauthor[name=Umair Waheed, color=orange]{UW}

\usepackage{etoolbox}
\newtoggle{chfinal}
\makeatletter
\@ifpackagewith{changes}{final}
  {\toggletrue{chfinal}}
  {\togglefalse{chfinal}}
\makeatother

\usepackage[normalem]{ulem}
\usepackage{environ}

\NewEnviron{chaddedequation}
{%
  \iftoggle{chfinal}
    {\begin{equation}\BODY\end{equation}}
    {\begin{equation}\color{blue}\BODY\end{equation}}%
}

\NewEnviron{chaddedequation*}
{%
  \iftoggle{chfinal}
    {\begin{equation*}\BODY\end{equation*}}
    {\begin{equation*}\color{blue}\BODY\end{equation*}}%
}

\NewEnviron{chdeletedequation}
{%
  \iftoggle{chfinal}
    {}
    {%
      \begin{equation}
      \color{blue}\mbox{\sout{$\displaystyle \BODY$}}
      \end{equation}
    }%
}

\NewEnviron{chdeletedequation*}
{%
  \iftoggle{chfinal}
    {}
    {%
      \begin{equation*}
      \color{blue}\mbox{\sout{$\displaystyle \BODY$}}
      \end{equation*}
    }%
}

\NewEnviron{chaddedtable}[1][htbp]
{%
  \iftoggle{chfinal}
    {\begin{table}[#1]\BODY\end{table}}
    {\begin{table}[#1]\color{blue}\BODY\end{table}}%
}
\NewEnviron{chaddedfigure}[1][htbp]
{%
  \iftoggle{chfinal}
    {\begin{figure}[#1]\BODY\end{figure}}
    {\begin{figure}[#1]\color{blue}\BODY\end{figure}}%
}

\journal{Artificial Intelligence in Geosciences}

\begin{document}

\begin{frontmatter}



\title{Explainable AI for microseismic event detection} 


\author[1]{Ayrat Abdullin}
\author[2]{Denis Anikiev}
\author[1,2]{Umair Bin Waheed}

\affiliation[1]{organization={Department of Geosciences, King Fahd University of Petroleum and Minerals},
            city={Dhahran},
            postcode={31261}, 
            country={Saudi Arabia}}

\affiliation[2]{organization={Center for Integrative Petroleum Research, 
                            King Fahd University of Petroleum and Minerals},
            city={Dhahran},
            postcode={31261}, 
            country={Saudi Arabia}}

\begin{abstract}

Deep neural networks like PhaseNet show high accuracy in detecting microseismic events, but their black-box nature is a concern in critical applications. We apply \chreplaced{Explainable Artificial Intelligence}{explainable AI} (XAI) techniques, such as Gradient-weighted Class Activation Mapping (Grad-CAM) and Shapley Additive Explanations (SHAP), to interpret the PhaseNet model's decisions and improve its reliability. Grad-CAM highlights that the network's attention aligns with P- and S-wave arrivals. SHAP values quantify feature contributions, confirming that vertical-component amplitudes drive P-phase picks while horizontal components dominate S-phase picks, consistent with geophysical principles. Leveraging these insights, we introduce a SHAP-gated inference scheme that combines the model's output with an explanation-based metric to reduce errors. On a test set of 9,000 waveforms, the SHAP-gated model achieved an F1-score of 0.98 (precision 0.99, recall 0.97), outperforming the baseline PhaseNet (F1-score 0.97) and demonstrating enhanced robustness to noise. These results show that XAI can not only interpret deep learning models but also directly enhance their performance, providing a template for building trust in automated seismic detectors.
\chadded{The implementation and scripts used in this study will be publicly available at https://github.com/ayratabd/xAI\_PhaseNet.}

\end{abstract}









\end{frontmatter}


\section{Introduction}
\label{intro}

Microseismic monitoring detects and picks seismic events of very small magnitude, and it has greatly benefited from deep learning models in recent years. For instance, phase-picking neural networks such as PhaseNet~\citep{zhu2019phasenet} and the Earthquake Transformer~\citep{mousavi2020earthquake} are capable of automatically detecting P- and S-wave arrivals in continuous data, significantly speeding up the process of event cataloging. Additionally, other architectures have demonstrated notable effectiveness in various applications, including seismic facies classification~\citep{noh2023explainable} and full-waveform inversion~\citep{edigbue2025explaining}. These models frequently demonstrate superior performance compared to traditional methods in terms of accuracy and sensitivity. Nonetheless, a prominent challenge is that these models function as “black boxes”, which complicates the ability of seismologists to comprehend or have confidence in their decisions~\citep{guo2023interpretable, trani2022deepquake}. In high-stakes geoscience contexts, such as monitoring induced seismicity for CO$_{2}$ storage or mining, the absence of interpretability presents significant challenges, as reliability and transparency are crucial.  

Recent studies have brought to light particular challenges related to interpretability concerning phase-picking networks. For instance, the output score from PhaseNet represents a probability ranging from 0 to 1 at each time sample, which does not consistently align with the actual confidence of a pick. \citet{park2024making} noted that the prediction scores for both true and false picks can be inconsistently high or low, making it hard to set a threshold that cleanly separates real events from noise. In other words, these models’ output probabilities “do not necessarily correspond with the reliability” of the detection. Moreover, as noted by \cite{myren2025evaluation}, even when models such as PhaseNet appear highly accurate, their performance can fluctuate due to stochastic training and data‐sampling variability, highlighting the need for evaluation frameworks that explicitly quantify model uncertainty alongside accuracy. Consequently, two challenges emerge: (1) domain experts find it difficult to assess the level of trust they can place in an automated pick, and (2) there is ambiguity regarding which characteristics of the waveform influenced the model's decision-making process. The lack of clarity surrounding this issue impedes the implementation of \chadded{Artificial Intelligence (}AI\chadded{)} models in regular seismic monitoring, as professionals are hesitant to respond to detections that are not fully comprehensible to them.   

Explainable \chreplaced{Artificial Intelligence}{AI} (XAI) techniques present a valuable approach to tackle these challenges by shedding light on the inner workings of black-box models. In the field of geosciences, XAI has been increasingly recognized as a valuable approach for validating model behavior in relation to established domain knowledge. For example, feature-attribution methods such as \chreplaced{Shapley Additive Explanations~(SHAP)}{SHAP} and \chreplaced{Local Interpretable Model-agnostic Explanations~(LIME)}{LIME} have been applied in various contexts, including seismic facies classification~\citep{saikia2019seismic, lubo2022quantifying, bedle2024application}, full-waveform inversion~\citep{edigbue2025explaining}, and earthquake spatial probability assessment~\citep{jena2023explainable}, to determine which input attributes influence the model’s predictions. These methods, frequently referred to as post-hoc, serve to elucidate a model's workings after it has been trained. They effectively \chreplaced{``peel back''}{"peel back"} the black box, revealing whether the features deemed important by the model correspond with geophysical intuition or established factors. 

In the area of seismic signal analysis, applications of XAI are just starting to surface, yet initial findings are promising. \cite{trani2022deepquake} pioneered the application of activation visualizations for a \chreplaced{one-dimensional convolutional neural network~(1D CNN)}{1D CNN} detector, superimposing filter outputs on the raw waveform to identify which time segments activated the response of the network. Their qualitative \chreplaced{``heatmaps''}{"heatmaps"} revealed that high-energy onset arrivals of P-waves significantly activated specific convolutional filters. \cite{bi2021explainable} introduced a refined \chreplaced{Gradient-weighted Class Activation Mapping~(Grad-CAM)}{Grad-CAM} approach tailored for time-series data, known as \chreplaced{Explainable Upsampling Gradient-weighted Class Activation Mapping~(EUG-CAM)}{EUG-CAM}. This method projects the learned features of a CNN back onto the time-frequency domain. Through the process of upsampling the activations from the final convolutional layer, the researchers generated high-resolution explanation plots. These plots notably illustrated a surge of high-frequency energy coinciding with the arrival of the P-wave, which emerged as a critical characteristic for the classification of a microseismic event. Saliency-based methods have shown that deep networks tend to concentrate on seismic characteristics that are recognizable to humans, even in the absence of explicit instructions. 

In addition to visual heatmaps, other researchers have utilized Layer-wise Relevance Propagation (LRP) and similar techniques to explore waveform classifiers. \cite{majstorovic2023interpreting} applied LRP to a single-station earthquake detector CNN and could thus trace which parts of the input contributed most to a detection. They found that the CNN had in fact learned to recognize where an earthquake’s signal is within a long window (something not given during training) and that many of the network’s salient features corresponded to physical aspects of the signal, such as the P-wave and S-wave portions and their frequency content. Notably, their analysis uncovered distinctions between the strategy employed by the CNN and that of a human analyst or a traditional STA/LTA trigger. This highlights that the model occasionally relies on more nuanced features that may not be apparent through visual inspection. In a similar vein, \cite{jiang2024explainable} utilized LRP on a microseismic classification model to analyze both accurate and inaccurate choices. In the context of true events, LRP verified that the network was focused on relevant waveform characteristics, such as a sudden increase in amplitude indicating an arrival, while for noise, it assisted in diagnosing failure modes. For instance, a false positive where the model was “fooled” by a transient noise spike, or a missed event where the signal lacked the frequency characteristics the model expected. These studies demonstrate how post-hoc explanations can expose the question of whether a model’s “reasoning” corresponds with geophysical reality and assist in identifying the reasons behind its misclassification of certain cases.  

Even with these advancements, a significant gap in the existing literature is the application of SHAP\chreplaced{}{ (Shapley additive explanations)} in deep seismic waveform models. SHAP represents a robust game-theoretic method that allocates an importance value to each feature for a specific prediction. Nonetheless, the direct application of SHAP to high-dimensional inputs, such as time series data, poses significant challenges. In microseismic monitoring, each waveform may contain thousands of sample points (features), which renders classical SHAP analysis both computationally demanding and challenging to interpret in its raw form. Consequently, to our knowledge, no prior work has reported using SHAP on a CNN-based microseismic event detector. Although SHAP has proven useful in interpreting models for various geophysical tasks, including full-waveform inversion~\citep{edigbue2025explaining}, seismic data denoising~\citep{antariksa2025xai}, and regional earthquake hazard assessment~\citep{jena2023explainable}, its use in the specific area of high-temporal-resolution waveform phase detection has yet to be investigated. 
Our objective is to address this gap by illustrating the ways in which SHAP can be tailored for time-series seismic data. By aggregating SHAP values meaningfully (for example, summing contributions over time segments or sensor components), we extract clear insights from PhaseNet’s internal decision logic.

In this article, we make two important contributions. 
First, we apply Grad-CAM and SHAP to PhaseNet to interpret its microseismic event detection behavior. PhaseNet is a widely used phase-picking model (originally developed for earthquake \chreplaced{P- and S-wave~(P/S)}{P/S} arrival timing) that we have adapted for microseismic binary (signal vs. noise) event detection. Using XAI, we reveal which parts of the waveform and which sensor components PhaseNet relies on for detecting events. 
Second, we go beyond interpretation by using the XAI results to enhance PhaseNet’s performance. We develop a simple yet effective SHAP-gated inference scheme that uses the explanation (SHAP values) to decide whether to accept or reject a detection. 
\chadded{Specifically, for each waveform window we compute SHAP contributions for the three components (East, North, and Vertical) (E, N, Z) for both the P- and S-phase outputs, take the mean absolute value of these six attributions, and use this quantity as an explanation-based evidence score. A detection is accepted when this score exceeds a threshold calibrated on the training set; otherwise it is rejected.}
By incorporating this scheme, we improve the precision and recall of PhaseNet on a real microseismic dataset.
To the best of our knowledge, this is the first instance in seismic event detection where explanations are used to inform the model’s output in \chreplaced{post-hoc decision fusion}{a closed-loop fashion}. This approach represents a developing trend in which XAI is utilized not just for interpretation but also for enhancing performance directly. This is illustrated in various fields, including the use of XAI for data augmentation in seismic denoising~\citep{antariksa2025xai}, predicting ground-motion parameters~\citep{Sun2023xai}, and in adjacent fields, e.g., for enhancing damage recognition accuracy in building damage detection~\citep{Wang2025xai}. 

\chreplaced{
Our findings demonstrate that this approach yields a more consistent and trustworthy detector: on the test set, the SHAP-gated scheme improved the F1-score (harmonic mean of precision and recall) from 0.97 to 0.98 and increased recall from 0.96 to 0.97, while also showing greater robustness under progressively stronger noise contamination. These gains are important for practical geophysical monitoring, where more reliable event screening can reduce missed detections, improve analyst confidence in automated triggers, and support safer deployment of AI-assisted decision-making systems. Although our study focuses on PhaseNet, in the Discussion section we consider how the same explainability-guided strategy could be extended to other emerging models, including Transformer-based detectors, and how XAI may help enable broader deployment of geophysical AI models.
}{
Our findings demonstrate that the approach we employed results in a detector that is both more consistent and trustworthy, effectively tackling the reliability concerns highlighted by previous researchers. Our study primarily examines PhaseNet; however, in the Discussion section, we explore the potential application of these explainability techniques to other emerging models, such as Transformer-based detectors, and consider how XAI can play a crucial role in the deployment of geophysical AI models.
}

\section{Materials and Methods}
\label{mat_met}

\subsection{Microseismic Dataset and PhaseNet Model}
\label{data_model}

\chreplaced{The dataset used in this study consists of triggered waveforms recorded during hydraulic fracturing operations in British Columbia, Canada. The acquisition geometry comprised nine three-component surface seismic sensors distributed over an area of approximately 100 km$^2$. The original recordings were sampled at 250 Hz and subsequently downsampled to 100 Hz, which is sufficient for the present study because the dominant signal frequency is below 50 Hz. The cataloged events are low-magnitude induced microseismic events ($0.5 < M < 2.5$) with manually picked P- and S-wave arrival times, occurring at an average depth of 2.1 km and ranging from 1.7 to 2.4 km. 
The dataset comprises labeled waveform windows, with a consistent length of 30 seconds each. 
The event windows include local events as well as events recorded at distances exceeding 10 km from the array centroid, while the noise windows were extracted from continuous recordings during intervals without detected seismicity before operations began and include field-specific non-seismic background and transient noise.
Each window is assigned a binary label: signal (event) if it contains a microseismic arrival, or noise if no event is present. 
We curated a balanced dataset of approximately 10,000 windows. For each run, we randomly selected 100 windows as a balanced training subset for threshold tuning, and then evaluated the model on a separate test set of 9,000 windows drawn from the remaining 9,900 windows.
}{Our evaluation of the explainable AI approach involves utilizing a dataset comprised of microseismic waveforms, which capture the characteristics of minor induced earthquakes, along with noise recordings. The dataset comprises labeled waveform windows, with a consistent length of 30 seconds each, obtained from continuous recordings in a practical monitoring setting. Each window is assigned a binary label: signal (event) if it contains a microseismic arrival, or noise if no event is present. We curated a balanced dataset of approximately 10,000 windows, split into a training set (100 windows) and a test set (9,000 windows). The signals in this dataset are low-magnitude seismic events (in the order of $0.5 < M < 2.5$) from a surface geophone array, with clear P-wave and S-wave onsets, superimposed on ambient noise. Examples of noise encompass instrument noise as well as a range of non-seismic transients that are characteristic of the field environment.}
All waveforms are preprocessed with amplitude normalization, which is standard for microseismic detection. We use three-component recordings (East-West, North-South, vertical) so that phase polarity differences can be leveraged by the model.
\chadded{For the harmonic-noise robustness experiments described later, the injected harmonic noise was taken from a separate field dataset acquired in Spain, consisting of five surface stations with strong pump-induced harmonic noise.}

\begin{figure}[!htbp]
\begin{center}
\includegraphics[width=0.99\textwidth]{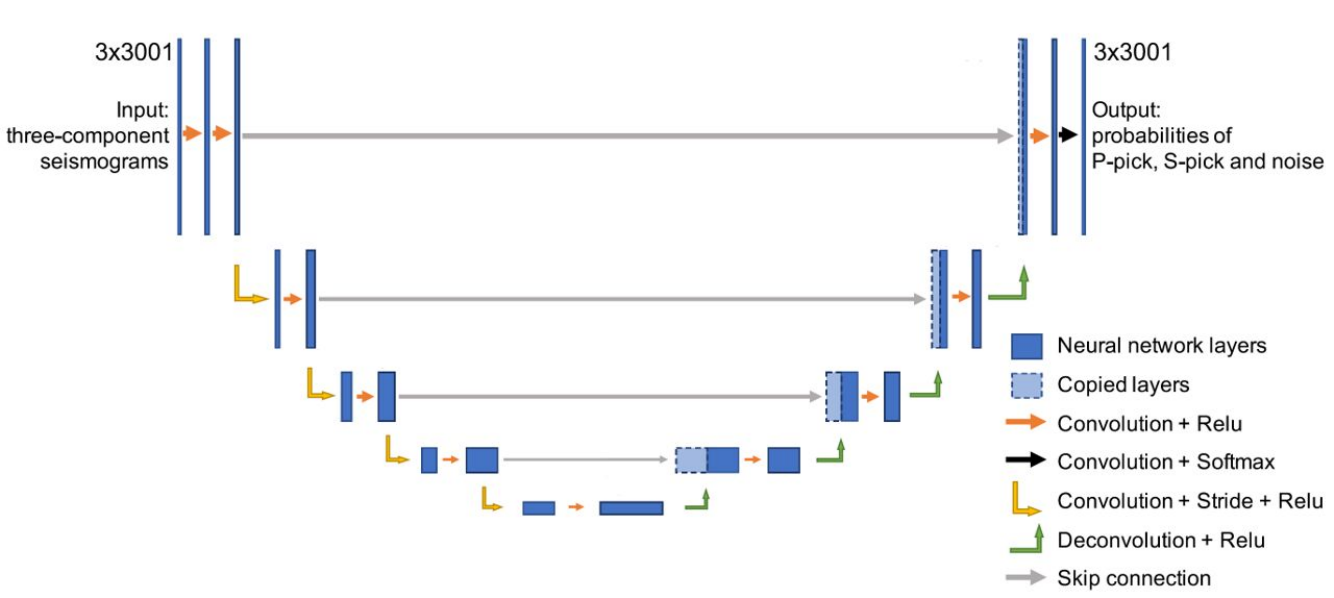}
\caption{
Schematic illustration of the network architecture. The input consists of 30-s three-component seismograms sampled at 100 Hz, yielding an input dimension of 3 × 3001. The network outputs three probability sequences of equal length, corresponding to P-pick, S-pick, and noise classes. Blue rectangles indicate neural network layers. Arrows denote operations, as summarized in the lower right corner. 
The model comprises four stages dedicated to down-sampling and four stages for up-sampling. Down-sampling is carried out using 1-D convolutions with a kernel size of 7 and a stride of 4. In contrast, up-sampling is performed through deconvolutions, which serve to restore the sequence length from the previous stage. Skip connections combine feature maps from each down-sampling stage with the corresponding up-sampling stage (indicated by dashed rectangles), thereby aiding in convergence. The last layer utilizes a softmax activation function to produce class probabilities (adapted from \cite{zhu2019phasenet}).
}
\label{fig:phasenet}
\end{center}
\end{figure}

Our base detection model is PhaseNet~\citep{zhu2019phasenet}, a deep convolutional neural network originally designed for picking P and S phase arrival times. PhaseNet’s architecture follows a U-Net style fully convolutional network with an encoder-decoder structure (Figure~\ref{fig:phasenet}). In our implementation, the model takes a multi-component waveform window as input and produces as output a set of probability traces – one for each class of interest (noise, P arrival, S arrival). For binary event detection, we interpret the PhaseNet output as a single probability of “event present” within the window, derived from the maximum predicted likelihood of a P or S arrival in that window. Essentially, if PhaseNet produces a class probability that exceeds a chosen threshold for either the P or S channel within the window, the window is classified as containing an event. We tested the PhaseNet on our microseismic dataset using supervised learning: windows with actual events were labeled positive, and noise-only windows were negative. The pre-trained PhaseNet achieved ~97\% classification accuracy on the held-out test set, corresponding to high initial precision and recall (details in Section~\ref{results}). However, like prior studies, we observed that setting an optimal decision threshold on the PhaseNet output was non-trivial. A simple 0.5 probability cutoff was not satisfying, and tuning the threshold involved trading off false negatives versus false positives. This observation motivated us to investigate the incorporation of explainability metrics into the decision process.

\subsection{Grad-CAM for Waveform Data}
\label{gradcam}

To interpret which parts of a waveform influence PhaseNet’s predictions, we applied the Gradient-weighted Class Activation Mapping (Grad-CAM) technique~\citep{selvaraju2017grad}. Grad-CAM is a general method that produces a coarse “heatmap” of importance by using the gradients of the target class score with respect to the convolutional layers of the network. In image applications, Grad-CAM highlights image regions most responsible for a classification; here, we adapt it to 1D time-series data. We followed the procedure of \cite{selvaraju2017grad} in the context of our 1D CNN: we fed a waveform through PhaseNet and obtained the event probability output. We then computed the gradient of that output (for a given window) with respect to the feature maps of the final convolutional layer. When these gradients are averaged globally across all time positions, they provide weights that reflect the significance of each filter's activation in relation to the event prediction. Next, we took a weighted linear combination of the feature maps (of the final convolutional layer) using these gradient-derived weights, and then applied a ReLU (rectified linear unit) to keep only positive influences. The result is a coarse activation map across the time dimension, which we then linearly interpolated to the original waveform length to obtain a time series of “importance scores” -- the Grad-CAM heatmap. It is essential to recognize that by utilizing the final convolutional layer, Grad-CAM emphasizes high-level semantic features, albeit at the expense of spatial resolution. Researchers have observed that this may obscure fine-grained details and have suggested alternative strategies, including optimal layer selection~\citep{yoo2022vibration} and the fusion of heatmaps from multiple layers~\citep{li2023multilayer}, to tackle this trade-off. In the context of this research, the conventional Grad-CAM method was considered adequate for identifying the main P- and S-wave energy packets.

In practice, we generated Grad-CAM explanations for many test examples, both true events and noise. 
The Grad-CAM output for each waveform was then overlain on the waveform plot for visualization. This allowed us to see, for example, if the network was focusing on the P-wave onset, the S-wave arrival, or perhaps some noise burst. 
It should be noted that standard Grad-CAM can miss features that have a negative influence on the prediction (since the ReLU truncates negative gradients). However, since we are primarily interested in what supports an event detection (the positive evidence), this was acceptable. For completeness, one could use guided backpropagation or LRP to capture inhibitory factors, but that was beyond our scope. Our Grad-CAM implementation yields an approximate explanation of where PhaseNet ``looks'' in time to decide if a window contains an event.

\subsection{SHAP Value Analysis}
\label{shap}

While Grad-CAM provides a visual localization of important regions, it does not quantify the contribution of each input feature. We therefore turned to Shapley additive explanations (SHAP; ~\cite{lundberg2017unified}) to attribute an importance value to every sample in the waveform. SHAP interprets the prediction of a model by computing the contribution of each feature (input dimension) toward the difference between the model’s output and a baseline output. Intuitively, a positive SHAP value for a given sample (at a specific time and component) means that the sample increased the model’s confidence in the event class, whereas a negative value means it pushed the model toward predicting noise.

Directly computing Shapley values for every sample in a long seismic waveform is intractable, so we simplified the problem by focusing on component-level attributions. For each waveform window, we generated all possible combinations of masked components (E, N, Z) with either the true signal or a baseline replacement (\chreplaced{zeros}{zero or mean}). This yields the full set of $2^3 = 8$ coalition values, from which exact Shapley contributions can be computed without approximation. For each subset, we evaluated PhaseNet’s detection score (e.g., maximum P or S probability in the window) and then applied the Shapley value formulas to estimate the marginal contribution of each component. The resulting importance rankings $\phi_E$, $\phi_N$, $\phi_Z$ for P and S channels quantify how much each component contributes to the detection. 
This component-masking approach thus provides interpretable, mathematically grounded attributions while remaining computationally efficient.

Let $\mathbf{X}\in\mathbb{R}^{3\times L}$ be a single three–component window (E, N, Z) with $L=3001$ samples.

For a coalition mask $\mathbf{m}=(m_E,m_N,m_Z)\in\{0,1\}^3$, we form
\[
\mathbf{X}^{(\mathbf{m})}=\mathbf{m}\odot\mathbf{X},
\]
where $\odot$ denotes broadcasting and element-wise multiplication. The dropped channels are replaced by zeros (baseline).

Let $f(\cdot)$ be PhaseNet’s softmax output and $c\in\{\text{N},\text{P},\text{S}\}$ the target class index. The score of a (possibly masked) window is
\[
V_{\mathbf{m}}
=\; v\!\left(\mathbf{X}^{(\mathbf{m})};\,c\right)
=\; \max_{1\le t\le L}\; f_c\!\left(\mathbf{X}^{(\mathbf{m})}\right)_t .
\]

With three channels, we evaluate the eight coalitions
\[
V_{000},\,V_{100},\,V_{010},\,V_{001},\,V_{110},\,V_{101},\,V_{011},\,V_{111},
\]
where, e.g., $V_{100}$ keeps only E, $V_{011}$ keeps N and Z, etc.

With $n=3$ features, the Shapley weight for a subset $S$ not containing $i$ is $w(|S|)=\frac{|S|!(n-|S|-1)!}{n!}$, i.e., $w(0)=\tfrac{1}{3}$, $w(1)=\tfrac{1}{6}$, $w(2)=\tfrac{1}{3}$. The channel attributions $\phi_E,\phi_N,\phi_Z$ for a single window are
\[
\boxed{
\begin{aligned}
\phi_E &= \tfrac{1}{3}\,(V_{100}-V_{000})
       + \tfrac{1}{6}\,(V_{110}-V_{010})
       + \tfrac{1}{6}\,(V_{101}-V_{001})
       + \tfrac{1}{3}\,(V_{111}-V_{011}),\\[3pt]
\phi_N &= \tfrac{1}{3}\,(V_{010}-V_{000})
       + \tfrac{1}{6}\,(V_{110}-V_{100})
       + \tfrac{1}{6}\,(V_{011}-V_{001})
       + \tfrac{1}{3}\,(V_{111}-V_{101}),\\[3pt]
\phi_Z &= \tfrac{1}{3}\,(V_{001}-V_{000})
       + \tfrac{1}{6}\,(V_{101}-V_{100})
       + \tfrac{1}{6}\,(V_{011}-V_{010})
       + \tfrac{1}{3}\,(V_{111}-V_{110}) .
\end{aligned}}
\]

In a general case, for a batch of $N$ windows, we report per-channel importance as the mean absolute Shapley value:
\[
\mathrm{Imp}_j \;=\; \frac{1}{N}\sum_{n=1}^N \bigl|\phi_j^{(n)}\bigr|,\qquad j\in\{E,N,Z\}.
\]

The SHAP component importance refers to the total contribution attributed to an entire component of the sensor.  
We summarized the SHAP results by computing, for each class (event vs. noise), the average contribution of each component (E, N, Z). In addition, we recorded how often a given component provided the largest SHAP value, that is, how frequently that channel contributed the most to the model’s decision compared with the other two\chadded{ (reported in Table~\ref{tab:shap_summary} as ``\% Dominant'')}.
These metrics help relate the model’s behavior to the known seismic wave propagation characteristics. We also created SHAP summary plots where each dot represents a feature (\chreplaced{component)}{time sample) colored by component}, plotted against its SHAP value -- this visualizes the spread and magnitude of contributions for \chreplaced{signal and noise windows (Figure~\ref{fig:violin5K})}{signals and noises}.

It is worth noting that SHAP values offer a signed attribution -- some inputs can actually lower the event probability. In our case, however, we found that most features with significant magnitude had positive SHAP values for true events (they added to the likelihood of an event). Negative contributions were typically small and associated with scattered noise oscillations, which slightly push the model towards the “no-event” decision. For simplicity and interpretability, we focused on the positive SHAP contributions as indicators of features that support the presence of an event.

\chadded{To further understand the joint contributions of the components, we extended our framework to compute pairwise Shapley Interaction Indices. While individual Shapley values measure a component's marginal importance, the interaction index measures whether two components work synergistically (a positive value) or redundantly (a negative value). Because we already evaluate the full set of $2^3=8$ coalition values for the three components, the pairwise interaction for any two components~(e.g., E and N) can be calculated exactly by taking the difference between their combined marginal contribution and the sum of their individual marginal contributions.}

\subsection{SHAP-Gated Inference Scheme}
\label{shap_gate}

Beyond offline analysis, we integrated the explainability results into the PhaseNet’s decision logic. Our approach, termed SHAP-gated inference, uses a combination of SHAP values to classify a waveform as an event. The rationale comes from our observation that true events tend to produce not only a high model probability but also multiple significant SHAP contributions, whereas false positives often have only one weak transient indication of evidence (either in probability or SHAP).

\chadded{We therefore defined two scalar decision statistics for each window: (1) the PhaseNet event-probability score}

\begin{chaddedequation*}
P_{\max}=\max_{1\le t\le L}\max\left\{f_P(X)_t,\;f_S(X)_t\right\},
\end{chaddedequation*}

\chreplaced{
\noindent and (2) the SHAP evidence statistic \(S_6\), defined as the mean of the six absolute SHAP values (E, N, Z for both P- and S-wave probabilities) in that window. 
}{
We therefore defined two metrics for each window: (1) the PhaseNet output probability of an event (specifically, the maximum P-or-S probability in the window), and (2) the SHAP "mean6", defined as the mean of the 6 SHAP values (E, N, Z for both P- and S-wave probabilities) in that window.
}
Through exploratory analysis on the training set, we found that taking the mean of SHAP feature contributions gave a robust summary of the “amount of explanatory evidence” in a detection. 
Intuitively, a real event might trigger several strong features (e.g., P onset on Z component, S onset on horizontals, etc.), yielding \chreplaced{six}{6} high SHAP values whose mean is large. A spurious detection might only have one or two moderate features, and then the mean of six (including some zeros or low values) would be much lower.

Our decision rule is as follows: for each window, we compute six absolute Shapley values:

\[
\bigl\{\,|\phi_{\text{P},E}|,\,|\phi_{\text{P},N}|,\,|\phi_{\text{P},Z}|,\,    |\phi_{\text{S},E}|,\,|\phi_{\text{S},N}|,\,|\phi_{\text{S},Z}|\,\bigr\}.
\]

The decision statistic is their mean,

\[
S_6 = \frac{1}{6}\Bigl(
|\phi_{\text{P},E}| + |\phi_{\text{P},N}| + |\phi_{\text{P},Z}| +
|\phi_{\text{S},E}| + |\phi_{\text{S},N}| + |\phi_{\text{S},Z}|
\Bigr).
\]

\chadded{Using thresholds \(\tau_{\mathrm{PROB}}\) and \(\tau_{\mathrm{SHAP}}\), the probability- and SHAP-based decision rules are}

\begin{chaddedequation*}
\hat y_{\mathrm{PROB}}=\mathbf{1}\!\left[P_{\max}\ge \tau_{\mathrm{PROB}}\right],
\end{chaddedequation*}

\begin{chaddedequation*}
\hat y_{\mathrm{SHAP}}=\mathbf{1}\!\left[S_6\ge \tau_{\mathrm{SHAP}}\right].
\end{chaddedequation*}

\chdeleted{With a threshold $SHAP$, the classification rule is}

\begin{chdeletedequation*}
\hat{y} = \mathbf{1}\bigl[S_6 \;\ge\; SHAP\bigr],
\end{chdeletedequation*}

\chdeleted{i.e., the window is labeled as a signal if and only if the average SHAP importance across all six (P/S $\times$ components) exceeds the threshold.}

We optimized the thresholds \(\tau_{\mathrm{PROB}}\) and \(\tau_{\mathrm{SHAP}}\) \chreplaced{}{($PROB$, $SHAP$)} on the training set by sweeping values to maximize the F1-score\chdeleted{ (the harmonic mean of precision and recall)}. We then fixed these thresholds and applied the rule to the test set to evaluate the improvement in detection performance (see Section~\ref{results_detection}). It’s essential to note that the thresholds are dimensionless, but they are tied to our data normalization and model output scaling. In another setting, they would need recalibration. In effect, this SHAP-gating constitutes a simple post-hoc decision fusion, combining the model’s numeric output with an XAI-based feature metric.

\section{Results}
\label{results}

\subsection{Grad-CAM Reveals Model Focus on Seismic Phases}
\label{results_gradcam}

\begin{figure}[!htbp]
\begin{center}
\includegraphics[width=0.59\textwidth]{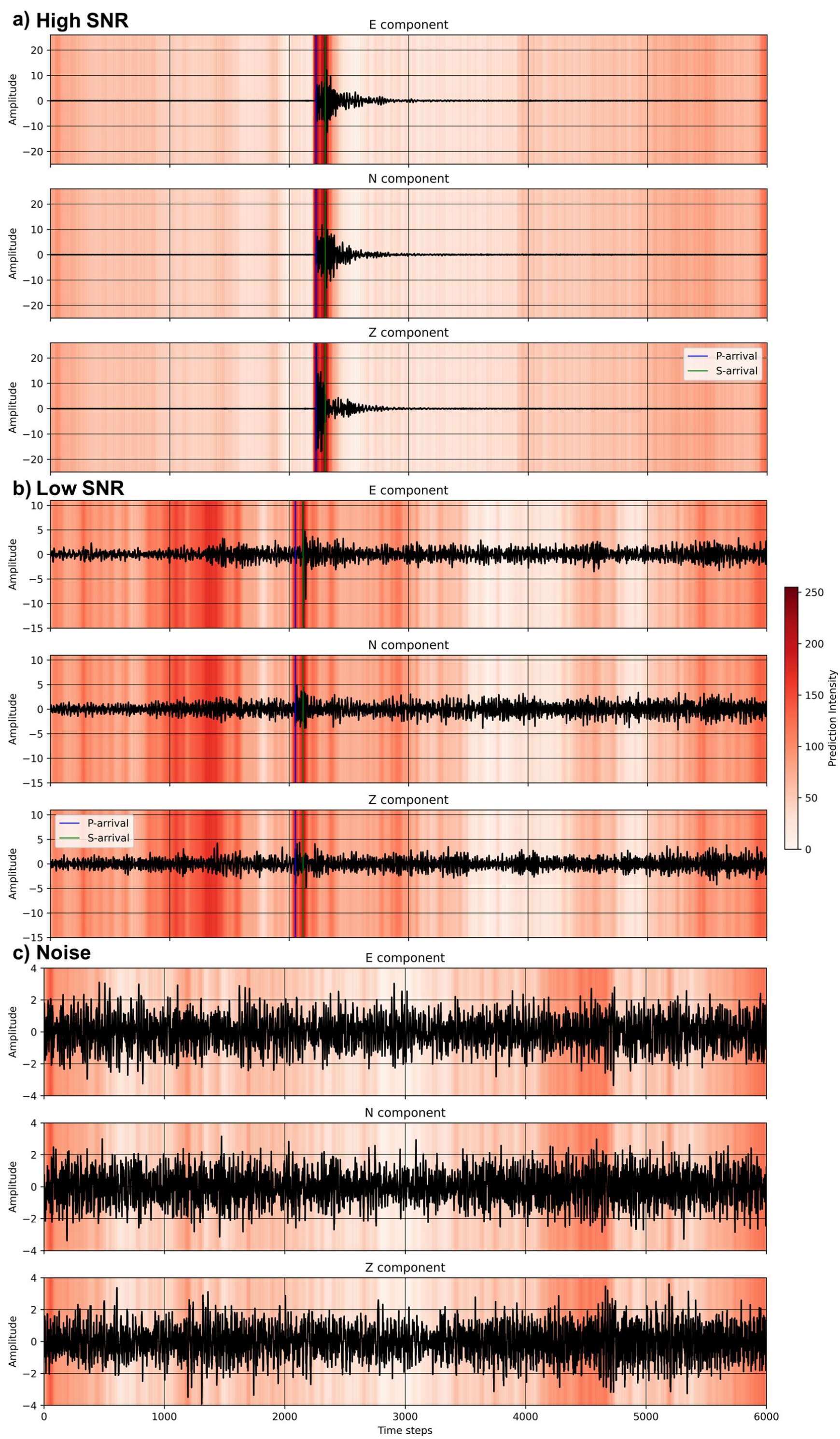}
\caption{
Grad-CAM visualizations of PhaseNet attention on the
\chreplaced{full three-component waveform record (E, N, and Z)}{vertical (Z) component} 
for three representative cases: (a) a high-SNR event with clear P- and S-wave arrivals, (b) a low-SNR event, and (c) a noise-only window. The heatmaps show Grad-CAM attribution intensity (red shading) overlaid on the waveform amplitude. Darker shades correspond to stronger model attention. For high-SNR events, attention is sharply concentrated around the P-wave onset with secondary activation near the S arrival; for low-SNR events, the attention remains aligned with the arrivals but becomes broader and less intense; for noise, attention is diffuse and unstructured, indicating no consistent phase-like focus.
}
\label{fig:gradcam3}
\end{center}
\end{figure}

Grad-CAM visualizations provided clear insights into which waveform segments PhaseNet relied on for event detection. Figure~\ref{fig:gradcam3} shows \chadded{three-component} examples for three representative cases: a \chreplaced{high signal-to-noise ratio~(SNR)}{high-SNR} event, a low-SNR event, and a pure noise window.

\begin{figure}[!htbp]
\begin{center}
\includegraphics[width=0.79\textwidth]{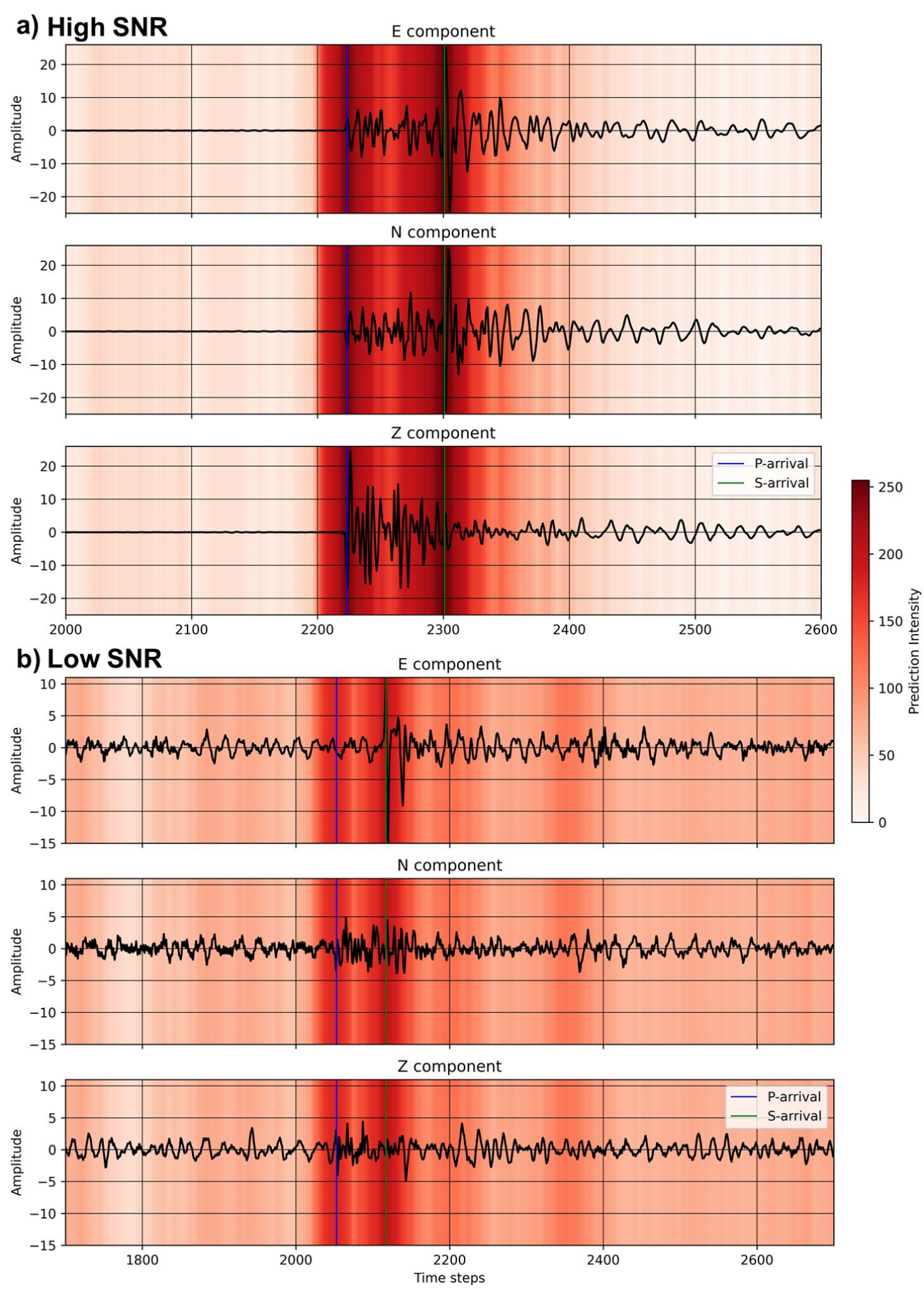}
\caption{
Zoomed-in Grad-CAM views of the 
\chreplaced{full three-component waveform record (E, N, and Z)}{vertical (Z) component}
for (a) a high-SNR and (b) a low-SNR event, corresponding to Figure~\ref{fig:gradcam3}a–b. The heatmaps emphasize the specific areas where the model exhibits the greatest emphasis in distinguishing between P- and S-waves arrivals.  The high-SNR example illustrates a distinct and clearly defined activation centered on the arrival of the P component, while, in the low-SNR scenario, there is a broader and less pronounced area of focus, which suggests a decrease in model confidence when faced with noisy conditions.
}
\label{fig:gradcam2}
\end{center}
\end{figure}

For the high-SNR event (Fig.~\ref{fig:gradcam3}a, \ref{fig:gradcam2}a), Grad-CAM activations are sharply concentrated around the manually picked P arrival. A secondary but weaker highlight is visible at the S arrival. The zoomed view (Fig.~\ref{fig:gradcam2}a) illustrates that the importance scores extend across several tens of samples around the onset, indicating that PhaseNet bases its decision not just on the very first sample but on the characteristic onset pattern. 

Importantly, outside of these arrival times, the Grad-CAM values are much smaller. Thus, the background coda and noise in the rest of the window do not strongly influence the model. This focus on real seismic phases gives us confidence that PhaseNet’s internal logic is \chreplaced{qualitatively consistent with seismic phase physics}{geologically/plausibly sound}, rather than latching onto unrelated artifacts. Similar observations have been reported by other authors using different explanation methods; for example, \cite{majstorovic2023interpreting} found that their CNN detector clearly “learned to recognize where the earthquake is within the sample window” via relevance mapping.

In the low-SNR event (Fig.~\ref{fig:gradcam3}b, \ref{fig:gradcam2}b), the attributions continue to correspond with the approximate P- and S-wave neighborhoods, but the responses are less sharply picked and more spread out than in the high-SNR case. This shows that the model still pays attention to the right parts of the waveform, even when there is more noise, but with less confidence and a wider time range.

In comparison, the noise-only example (Fig.~\ref{fig:gradcam3}c) shows a lack of coherent Grad-CAM focus. The activation is weak and scattered across the entire window without clustering near any particular onset. This pattern is consistent with a correct noise classification: the model doesn't find any phase-like features that would support an event label. 

These results show that PhaseNet's convolutional filters mostly focus on parts of the P and S arrivals that have physical meaning, even when the SNR changes. The lack of structured activations in noise windows reinforces the idea that the model is not just reacting to random spikes, but is also sensitive to real seismic phase patterns. These qualitative insights enhance confidence in PhaseNet's internal decision-making process and validate the subsequent quantitative attribution analysis.

\subsection{SHAP Highlights Key Features and Component Contributions}
\label{results_shap}

The SHAP analysis demonstrated distinct variations in the reliance of PhaseNet on each component for classifying P- and S-phases. Table~\ref{tab:shap_summary} shows the average absolute Shapley values with confidence intervals and the percentage of cases in which each component is the most important\chadded{ (frequently dominant)}.
\chreplaced{For P-class detections on signal windows, the vertical (Z) component has the biggest effect, with a mean contribution of 0.30 and a dominance frequency of 49.7\% (reported as ``\% Dominant'' in Table~\ref{tab:shap_summary}). This indicates that PhaseNet primarily uses vertical ground motion to identify P-wave arrivals. For S-class detections, the horizontal components (E and N) are more important, with mean absolute SHAP values of 0.36 and 0.33 and dominance frequencies of 50.5\% and 45.6\%, respectively. 
}{For P-class detections on signal windows, the vertical (Z) component has the biggest effect, with a mean contribution of 0.30 and being the most important in 50\% of cases. This proves that PhaseNet primarily uses vertical ground motion to find P-wave arrivals. For S-class detections, on the other hand, the horizontal components (E and N) are more important, with mean SHAP values of 0.36 and 0.33, and they are the most common in 46–51\% of cases.}
This is what seismologists expect, since S-wave energy is mostly recorded on horizontals. 

\begin{figure}[!htbp]
\begin{center}
\includegraphics[width=0.99\textwidth]{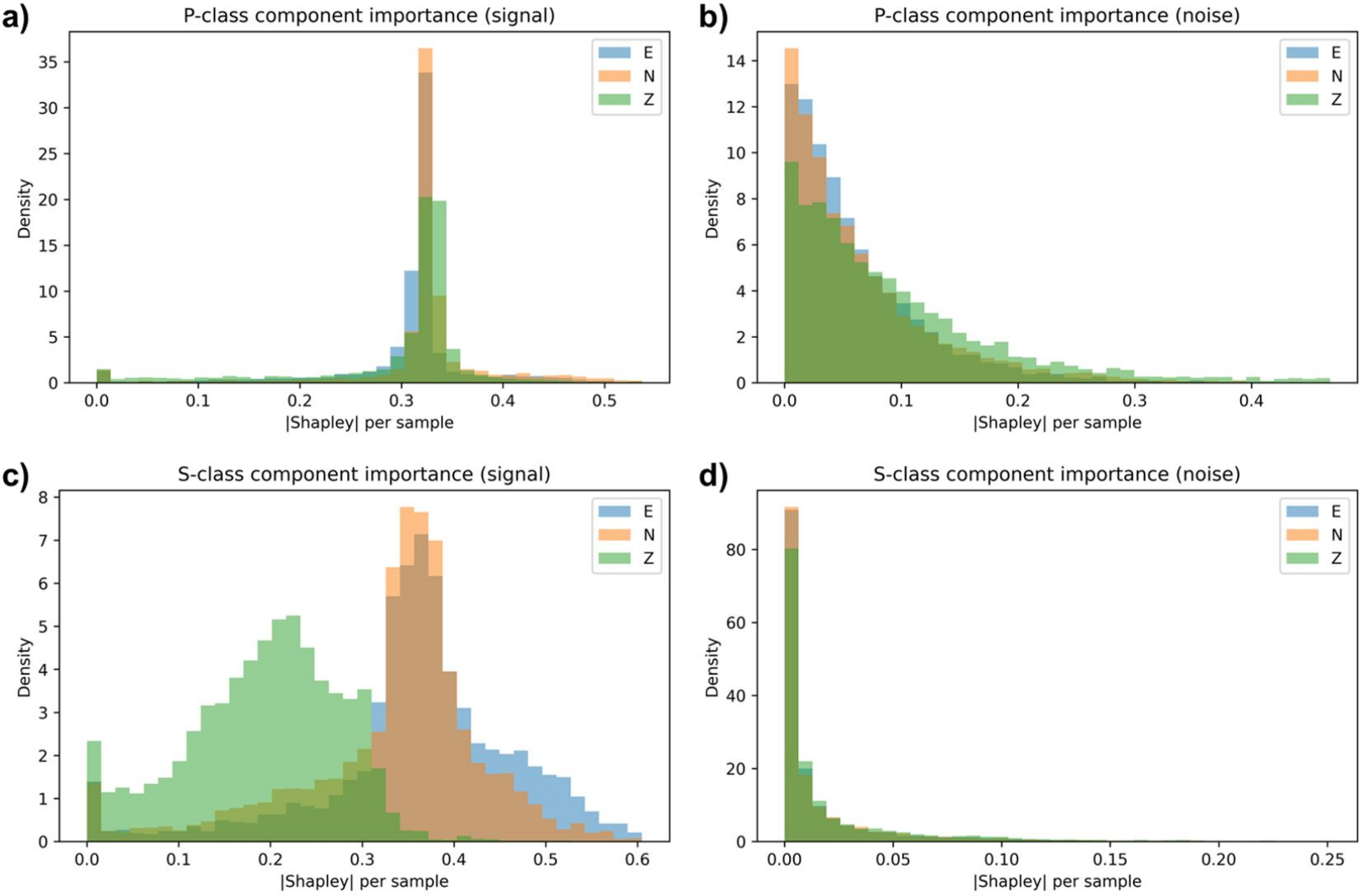}
\caption{
Distributions of absolute SHAP values ($|\phi|$) for 5,000 signal and 5,000 noise windows. Panels (a,b) show P-class attributions, and panels (c,d) show S-class attributions, separated by component (E, N, Z). 
\chreplaced{
For signal windows, the distributions are shifted toward higher $|\phi|$ values and exhibit component-specific structure: Z contributes most strongly to the P-class, whereas E and N dominate the S-class. In contrast, noise windows are concentrated near zero and overlap strongly across components, indicating the absence of a stable phase-consistent attribution pattern.
}{
For signal windows, vertical (Z) contributions slightly dominate the P-class, while horizontal components (E and N) dominate the S-class. Noise windows show SHAP magnitudes that are low across all components, which means that there is no clear evidence for either phase class. 
}
These distributions complement the summary statistics in Table~\ref{tab:shap_summary}.
}
\label{fig:hist5K}
\end{center}
\end{figure}

To further illustrate these component-level trends, Figure~\ref{fig:hist5K} shows the distributions of absolute SHAP values for predictions in the P- and S-class. 
\chreplaced{For signal windows (Fig.~\ref{fig:hist5K}a,c), the SHAP distributions are both stronger and more structured than for noise. In the P-class, the vertical component forms a narrow concentration at relatively high $|\phi|$ values, whereas in the S-class the horizontal E and N components are shifted toward larger magnitudes than Z. This pattern agrees with the component-level summary in Table~\ref{tab:shap_summary}, where signal windows show mean absolute SHAP values of about 0.30-0.33 for the P-class and 0.19-0.36 for the S-class, while the corresponding noise values remain much lower at about 0.06-0.10 and 0.02, respectively. Thus, the signal-to-noise separation in SHAP magnitude is substantial, indicating that true detections are supported by coherent attribution patterns rather than diffuse weak evidence.
For noise windows (Fig.~\ref{fig:hist5K}b,d), the histograms are concentrated near zero and overlap strongly across E, N, and Z, with no distinct component preference and no pronounced high-$|\phi|$ tail. Here, “no clear evidence” does not indicate under-sensitivity; rather, it indicates that the model does not identify a stable phase-consistent attribution signature in noise-only windows, which is the behavior expected from good discrimination. In visual terms, the tighter, higher-magnitude peaks in Fig.~\ref{fig:hist5K}a,c correspond to confident event-related feature usage, whereas the near-zero, overlapping distributions in Fig.~\ref{fig:hist5K}b,d correspond to the absence of persuasive evidence for either phase class. These distributions therefore reinforce that PhaseNet separates signal from noise using physically meaningful component patterns.}
{For signal cases (~\ref{fig:hist5K}a,c), SHAP values for the P-class cluster tightly around 0.25 on the Z component, whereas the S-class shows higher contributions on the horizontal E and N channels, consistent with shear-wave polarization. For noise windows (Fig.~\ref{fig:hist5K}b,d), on the other hand, all components show very low SHAP magnitudes ($< 0.1$) with no clear separation. This means that there is no coherent seismic-phase evidence. These histograms support the statistical patterns shown in Table~\ref{tab:shap_summary}, confirming that PhaseNet's attributions show real differences between the physics of P and S waves.}

\begin{figure}[!htbp]
\begin{center}
\includegraphics[width=0.99\textwidth]{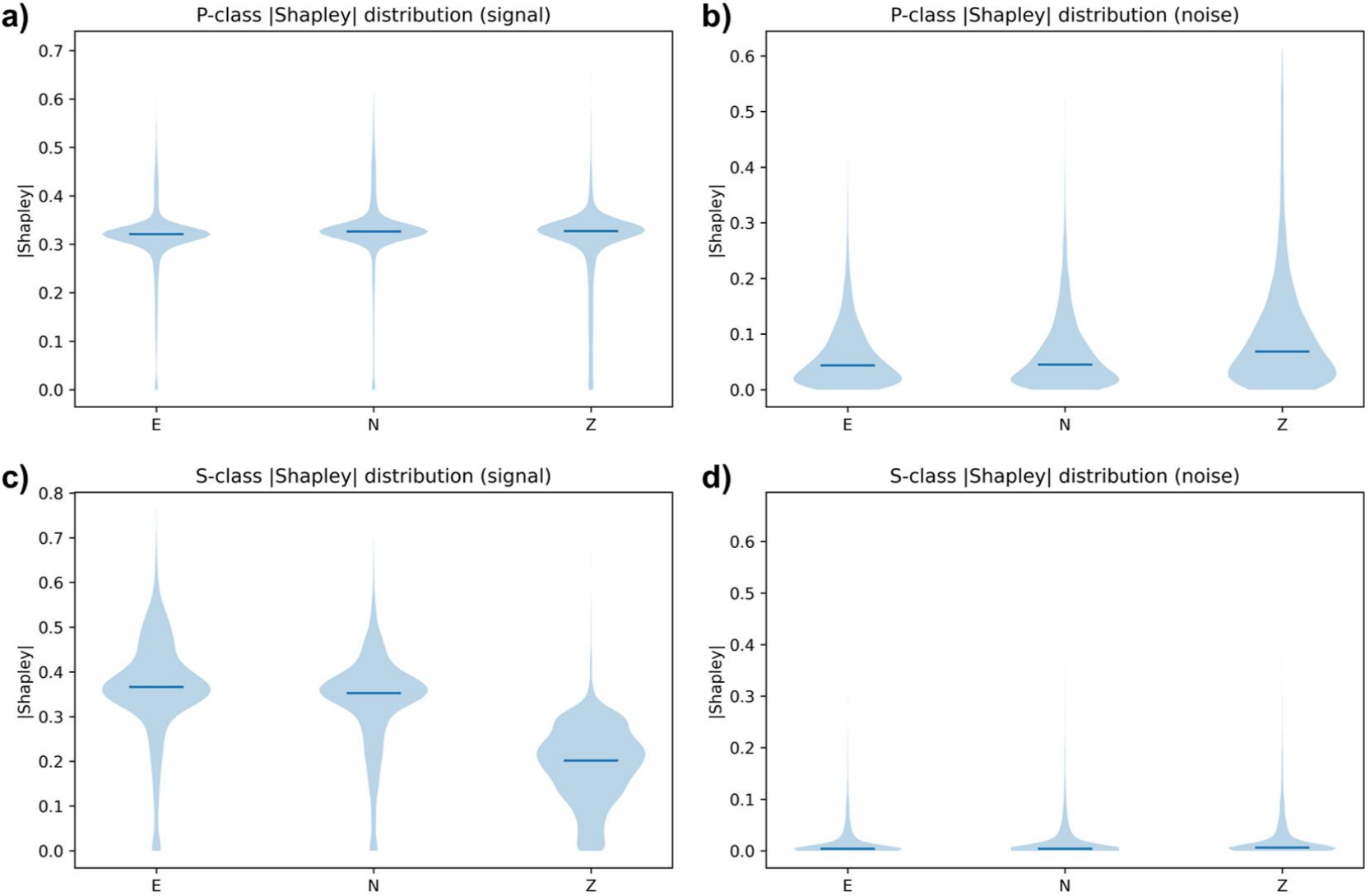}
\caption{
Violin plots of Shapley value distributions for 5,000 signal and 5,000 noise windows. (a) P-class SHAP distributions for signal windows; (b) P-class for noise; (c) S-class for signal; (d) S-class for noise. Signal windows show strong and coherent SHAP concentrations: the vertical (Z) component dominates P-phase detections, while horizontals (E and N) dominate S-phase detections. Noise windows exhibit uniformly low SHAP values across all components. Summary statistics are provided in Table~\ref{tab:shap_summary}.
}
\label{fig:violin5K}
\end{center}
\end{figure}

Figure~\ref{fig:violin5K} presents violin plots that effectively illustrate these trends. In the case of signal windows, the SHAP distributions exhibit a pronounced elevation centered around the arrivals of P- and S-waves (see Fig.~\ref{fig:violin5K}a,c), while for noise windows (Fig.~\ref{fig:violin5K}b,d) the values are much smaller and broadly distributed. In noise, mean SHAP values are only 0.06~(P-class) and 0.02~(S-class), confirming the lack of coherent explanatory evidence in the absence of true seismic phases. Notably, even though Z occasionally shows slightly higher noise attributions, these remain an order of magnitude weaker than for true events.

These findings collectively illustrate that PhaseNet’s attributions correspond with physical reality: Z predominates P-phase detections, while E and N prevail S-phase detections, and noise windows do not have any significant explanatory signal. This shows that PhaseNet not only attains a high level of accuracy, but it also uses features that have real-world meaning, which builds trust in its decision-making process. These observations led us to consider that by employing the model output to calculate the SHAP values, we can address certain misclassification errors. If we see a SHAP signature of an event, perhaps the detection should be accepted even if the raw probability was marginal. Likewise, if the model output is high but the SHAP evidence is not convincing, perhaps the detection should be discarded. This forms the basis of the SHAP-gated inference, the results of which we present next.

\begin{table}[!htbp]
\centering
\caption{Mean absolute SHAP values ($|\phi|$), 95\% confidence intervals (CI), and dominance percentages for each component (E, N, Z) across signal and noise windows. Values are reported separately for P-class and S-class predictions.}
\resizebox{\textwidth}{!}{%
\begin{tabular}{ccccccccc}
\hline
 & \multicolumn{4}{c}{P-class} & \multicolumn{4}{c}{S-class} \\
\cmidrule(lr){2-5} \cmidrule(lr){6-9}
Component & Mean $|\phi|$ & CI$_{lo}$ & CI$_{hi}$ & \% Dominant & Mean $|\phi|$ & CI$_{lo}$ & CI$_{hi}$ & \% Dominant \\
\hline
E (signal) & 0.31 & 0.31 & 0.31 & 15.1 & 0.36 & 0.36 & 0.37 & 50.5 \\
N (signal) & 0.33 & 0.32 & 0.33 & 35.2 & 0.33 & 0.33 & 0.34 & 45.6 \\
Z (signal) & 0.30 & 0.30 & 0.30 & 49.7 & 0.19 & 0.19 & 0.20 & 3.9 \\
\hline
E (noise) & 0.06 & 0.06 & 0.06 & 24.3 & 0.02 & 0.02 & 0.02 & 24.2 \\
N (noise) & 0.07 & 0.06 & 0.07 & 24.4 & 0.02 & 0.02 & 0.02 & 23.0 \\
Z (noise) & 0.10 & 0.10 & 0.10 & 51.4 & 0.02 & 0.02 & 0.02 & 52.8 \\
\hline
\end{tabular}}
\label{tab:shap_summary}
\end{table}

\subsection{\chadded{Joint Contributions and Reduced-Component Performance}}
\label{results_joint}

\chadded{
To answer whether the components provide synergistic information, particularly the horizontal pairs (N-E), we evaluated the pairwise Shapley Interaction Indices across the test set. The results (summarized in Table~\ref{tab:interactions} and Figure~\ref{fig:inter_hist}) reveal a strong predominance of redundancy over synergy. For both P-class and S-class predictions, the E-N pair exhibited negative interaction indices in over 80\% of the evaluated windows. This indicates that the horizontal components are highly redundant; detecting a signal on the N channel diminishes the marginal value of the E channel, as they capture overlapping physical information regarding phase arrivals. Interactions between vertical and horizontal components (E-Z, N-Z) also showed predominant redundancy, though to a slightly lesser extent ($\approx 60-70\%$), reflecting the distinct wavefield geometries they capture.
}

\begin{chaddedtable}[H]
\centering
\caption{Mean Shapley Interaction Indices, 95\% confidence intervals (CI), and percentages of synergistic ($>0$) versus redundant ($<0$) interactions for component pairs across the test set.}
\label{tab:interactions}
\resizebox{\textwidth}{!}{%
\begin{tabular}{lcccc}
\hline
\textbf{Pair} & \textbf{Mean} & \textbf{95\% CI ($CI_{lo}$, $CI_{hi}$)} & \textbf{\% Synergy ($>0$)} & \textbf{\% Redundancy ($<0$)} \\ \hline
\multicolumn{5}{c}{P-class Interactions} \\ \hline
E-N & -0.1233 & (-0.1257, -0.1209) & 18.9\% & 81.1\% \\
E-Z & -0.0870 & (-0.0894, -0.0845) & 30.7\% & 69.3\% \\
N-Z & -0.0951 & (-0.0977, -0.0925) & 28.2\% & 71.7\% \\ \hline
\multicolumn{5}{c}{S-class Interactions} \\ \hline
E-N & -0.1309 & (-0.1338, -0.1281) & 16.5\% & 83.5\% \\
E-Z & -0.0538 & (-0.0555, -0.0521) & 35.9\% & 64.1\% \\
N-Z & -0.0585 & (-0.0603, -0.0567) & 38.7\% & 61.3\% \\ \hline
\end{tabular}%
}
\end{chaddedtable}

\begin{chaddedfigure}[!htbp]
\begin{center}
\includegraphics[width=0.99\textwidth]{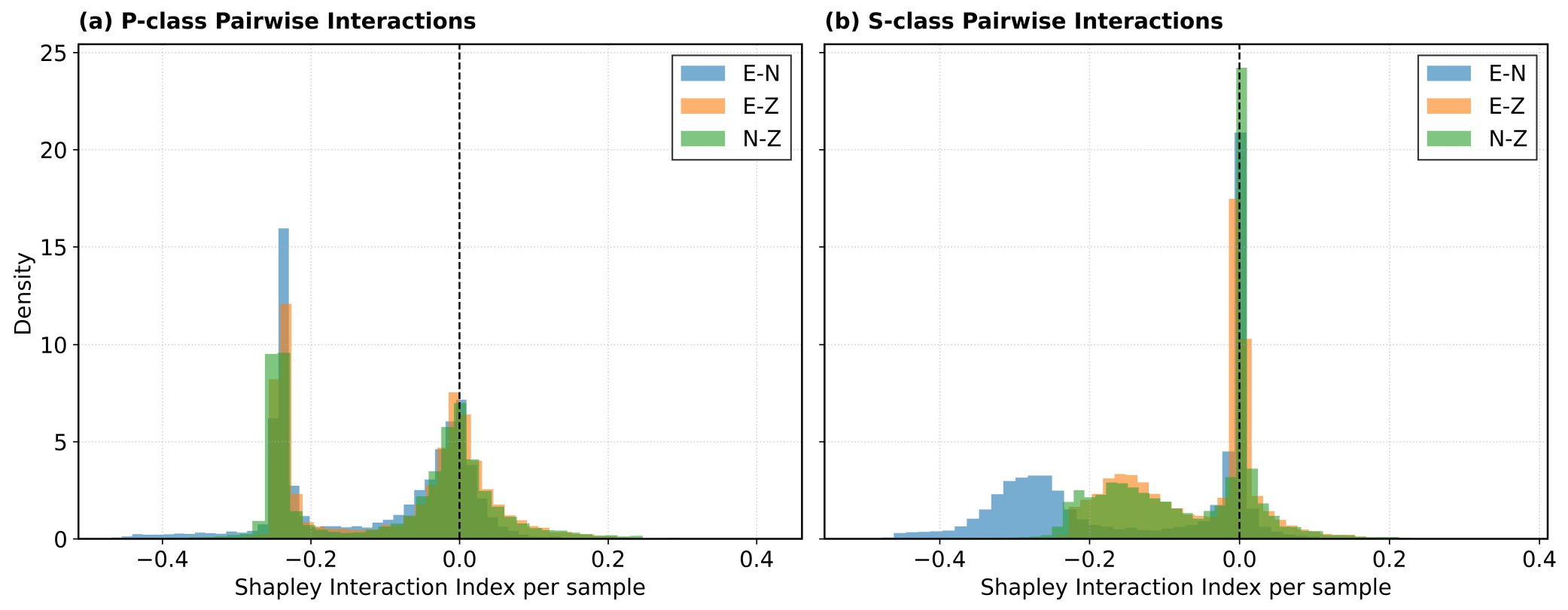}
\caption{
Distributions of pairwise Shapley Interaction Indices for P-class (left) and S-class (right) predictions. The dashed vertical line indicates zero (purely additive contributions). Values to the right indicate synergy, while values to the left indicate redundancy. The horizontal pair (E-N, blue) shows the strongest negative skew, indicating high redundancy between the horizontal components for both phase types.}
\label{fig:inter_hist}
\end{center}
\end{chaddedfigure}

\chreplaced{
Motivated by this high redundancy, we investigated whether a reduced-component system could achieve comparable performance to the full three-component (3C) system. We conducted a masked-input ablation study on the test set, systematically masking specific channels with zeros and re-evaluating the model's F1-score. As shown in Table~\ref{tab:ablation} and Figure~\ref{fig:abl_bar}, 2-component systems achieve highly comparable performance to the full 3C baseline. For instance, the baseline 3C system achieved a mean F1-score of 0.964. Dropping one horizontal channel to simulate a two-component (2C) system yielded F1-scores of 0.966 (E-Z) and 0.955 (N-Z). This physical ablation corroborates our SHAP interaction analysis: because the horizontal components are highly redundant, the model can maintain its predictive accuracy even when one is removed. However, dropping to a single vertical component (1C-Z) resulted in a significant performance drop (F1-score 0.882), emphasizing that at least one horizontal component is critical for accurate S-wave detection.
}{}

\begin{chaddedtable}[ht]
\centering
\caption{
\chreplaced{
Model detection performance (Mean F1 Score $\pm$ standard deviation across five cross-validation splits) for different simulated component configurations using a masked-input ablation study.}{}
}
\label{tab:ablation}
\begin{tabular}{lc}
\hline
\textbf{Configuration} & \textbf{F1 Score} \\ \hline
3C (E, N, Z) Baseline & $0.964 \pm 0.009$ \\ \hline
2C (E, N) & $0.972 \pm 0.002$ \\
2C (E, Z) & $0.966 \pm 0.008$ \\
2C (N, Z) & $0.955 \pm 0.004$ \\ \hline
1C (N) & $0.947 \pm 0.001$ \\
1C (E) & $0.945 \pm 0.007$ \\
1C (Z) & $0.882 \pm 0.017$ \\ \hline
\end{tabular}
\end{chaddedtable}

\begin{chaddedfigure}[!htbp]
\begin{center}
\includegraphics[width=0.99\textwidth]{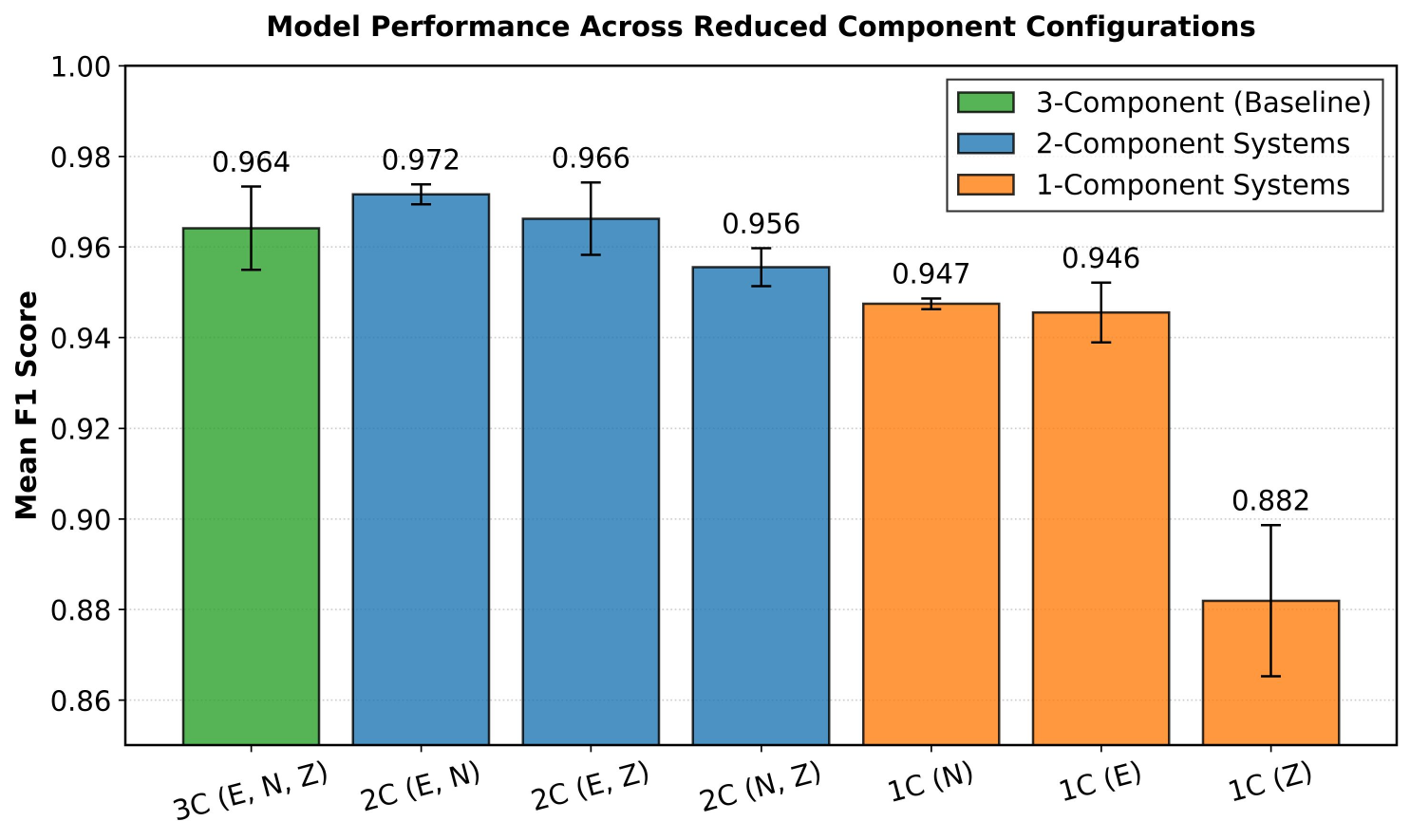}
\caption{Mean F1-scores for the PhaseNet model evaluated under various reduced-component configurations over five cross-validation splits. Two-component (2C) systems containing at least one horizontal channel (e.g., E-Z, N-Z) maintain performance highly comparable to the baseline 3-component (3C) system, corroborating the redundancy observed in the SHAP interaction analysis. Error bars represent one standard deviation.}
\label{fig:abl_bar}
\end{center}
\end{chaddedfigure}

\subsection{\chdeleted{3.3. }Improved Detection Performance with SHAP-Gated Inference}
\label{results_detection}

\chadded{
To explicitly link the reliability of decision-making to waveform quality near the detection threshold, we examined the relationship between the Signal-to-Noise Ratio (SNR) and the dispersion of attribution evidence. We quantify this attribution confidence using the standard deviation of the absolute SHAP values across the six phase-component combinations, defined as the SHAP dispersion ($D_{\mathrm{SHAP}}$):}
\begin{chaddedequation*}
D_{\mathrm{SHAP}} = \sqrt{\frac{1}{6}\sum_{k=1}^{6}\left(|\phi_k|-S_6\right)^2},
\end{chaddedequation*}
\chadded{where $S_6$ is the mean absolute SHAP value across the components. Figure~\ref{fig:shap_vs_snr} illustrates this relationship for 5,000 true seismic events and 5,000 pure noise windows. For high-SNR events, the model's attention is highly concentrated on specific phases and components, yielding low dispersion. However, as the SNR decreases toward the detection threshold (approaching 0 dB), the raw model score becomes less reliable, and the SHAP evidence becomes significantly more diffuse (higher $D_{\mathrm{SHAP}}$), visually mingling with the unstable attributions characteristic of pure noise. This demonstrates that near the decision boundary, isolated probability spikes are often driven by scattered, incoherent features. Consequently, we introduce a SHAP-gated inference scheme designed to filter out these unstable predictions by requiring coherent, multi-component evidence.}

\begin{chaddedfigure}[!htbp]
\begin{center}
\includegraphics[width=0.99\textwidth]{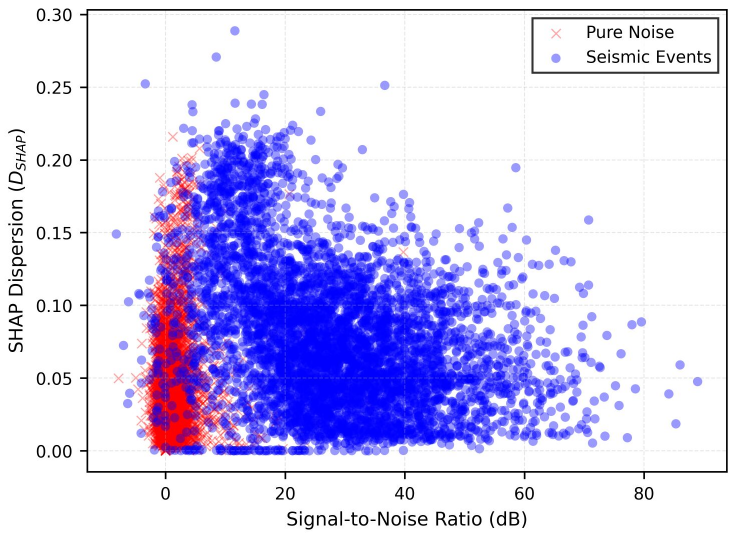}
\caption{Relationship between Signal-to-Noise Ratio (SNR) and SHAP attribution dispersion ($D_{\mathrm{SHAP}}$) for 5,000 seismic events (blue circles) and 5,000 pure noise windows (red crosses). At high SNR regimes, evidence is highly concentrated on physical seismic phases, resulting in low dispersion. As SNR decreases toward 0 dB, the attribution evidence becomes increasingly scattered and unstable, mirroring the behavior of pure noise. This highlights the necessity of utilizing SHAP evidence to evaluate decision-making reliability near the detection threshold.}
\label{fig:shap_vs_snr}
\end{center}
\end{chaddedfigure}

The central practical outcome of this study is that incorporating SHAP-based criteria into PhaseNet’s decision process improves its reliability, not only under clean conditions but also as noise levels increase. The optimal values for the training set of 100 samples (50 signals and 50 noise) were found to be \(\tau_{\mathrm{PROB}}\) \chdeleted{$PROB$} = 0.87 and \(\tau_{\mathrm{SHAP}}\) \chdeleted{$SHAP_{mean6}$} = 0.18. This means PhaseNet alone had to be >0.87 confident to trigger an event on probability alone; \chreplaced{in our algorithm}{otherwise}, we required a substantial SHAP evidence mean of 0.18 (in normalized units) to accept the event. We then evaluated the performance of the baseline model versus the SHAP-gated model on a balanced test set of 9,000 windows (4,500 true events and 4,500 noise).

On the clean dataset, the baseline PhaseNet (probability-only thresholding at 0.87) achieved an F1 score of 0.97, with a precision of 0.99 and a recall of 0.96. This corresponds to 186 false negatives and 45 false positives. With the SHAP-gated rule \(\tau_{\mathrm{SHAP}}\) \chdeleted{$SHAP_{mean6}$} = 0.18), performance improved to an F1 of 0.98, with Precision = 0.99 and Recall = 0.97, reducing false negatives to 140. These improvements were achieved without retraining the network -- simply by augmenting the decision rule with SHAP evidence.

\chadded{To ensure that this performance improvement was not an artifact of threshold variance on a limited tuning set, we conducted a repeated random sub-sampling validation (Monte Carlo cross-validation) on the clean dataset. We executed 50 independent splits, randomly re-sampling the 100-sample threshold-tuning set and the 9,000-sample test set. As detailed in Table~\ref{tab:cv_clean_stats}, the optimal \(\tau_{\mathrm{SHAP}}\) proved highly stable ($0.18 \pm 0.03$). Furthermore, the SHAP-gated inference outperformed the probability-only baseline in 86.0\% of the cross-validation splits, yielding a consistent F1 advantage. This confirms that while the absolute performance gain on the high-quality clean dataset is relatively modest, as the baseline is already approaching the performance ceiling, the explanation-based gating mechanism provides a statistically robust enhancement rather than a stochastic fluctuation.}

\begin{chaddedtable}[htpb]
\centering
\caption{
\chreplaced{
Statistical stability of detection performance over 50 Monte Carlo cross-validation splits on the clean dataset. Results are reported as Mean $\pm$ Standard Deviation. The SHAP-gated scheme demonstrates high threshold stability and consistently outperforms the probability-only baseline.
}{}
}
\label{tab:cv_clean_stats}
\resizebox{\textwidth}{!}{%
\begin{tabular}{lcc}
\hline
\textbf{Metric} & \textbf{Probability-Only Baseline} & \textbf{SHAP-Gated Inference} \\ \hline
Optimal Threshold & $0.76 \pm 0.11$ & $0.18 \pm 0.03$ \\
F1 Score & $0.967 \pm 0.0096$ & $0.973 \pm 0.0075$ \\
Win Rate & 14.0\% & \textbf{86.0\%} \\ 
\hline
\end{tabular}%
}
\end{chaddedtable}

Beyond clean conditions, we systematically tested robustness against increasing noise relative amplitude using both harmonic and random noise injection. For each relative amplitude, we ran five random cross-validation splits with a 100-sample balanced training set (50 signal, 50 noise) used to tune thresholds, and a separate 9,000-sample balanced test set for evaluation. This strict separation ensured that threshold optimization was not biased by the evaluation set.

\chadded{Here, the relative noise amplitude $a_{\mathrm{rel}}$ is defined as the ratio between the root-mean-square (RMS) amplitude of the injected noise and the RMS amplitude of the original signal. For each waveform component $c \in \{E,N,Z\}$, we compute}

\begin{chaddedequation*}
\mathrm{RMS}_{\mathrm{signal},c}
=
\sqrt{\frac{1}{T}\sum_{t=1}^{T} x_c(t)^2},
\end{chaddedequation*}

\chreplaced{and scale the injected noise $n_c(t)$ such that}{}

\begin{chaddedequation*}
\mathrm{RMS}_{\mathrm{noise},c}
=
a_{\mathrm{rel}}\,\mathrm{RMS}_{\mathrm{signal},c}.
\end{chaddedequation*}

\chadded{Thus, ``relative amplitude'' in this study refers to an RMS ratio, applied independently to each component, rather than a peak-to-peak or maximum-amplitude ratio.}

\begin{figure}[!htbp]
\begin{center}
\includegraphics[width=0.99\textwidth]{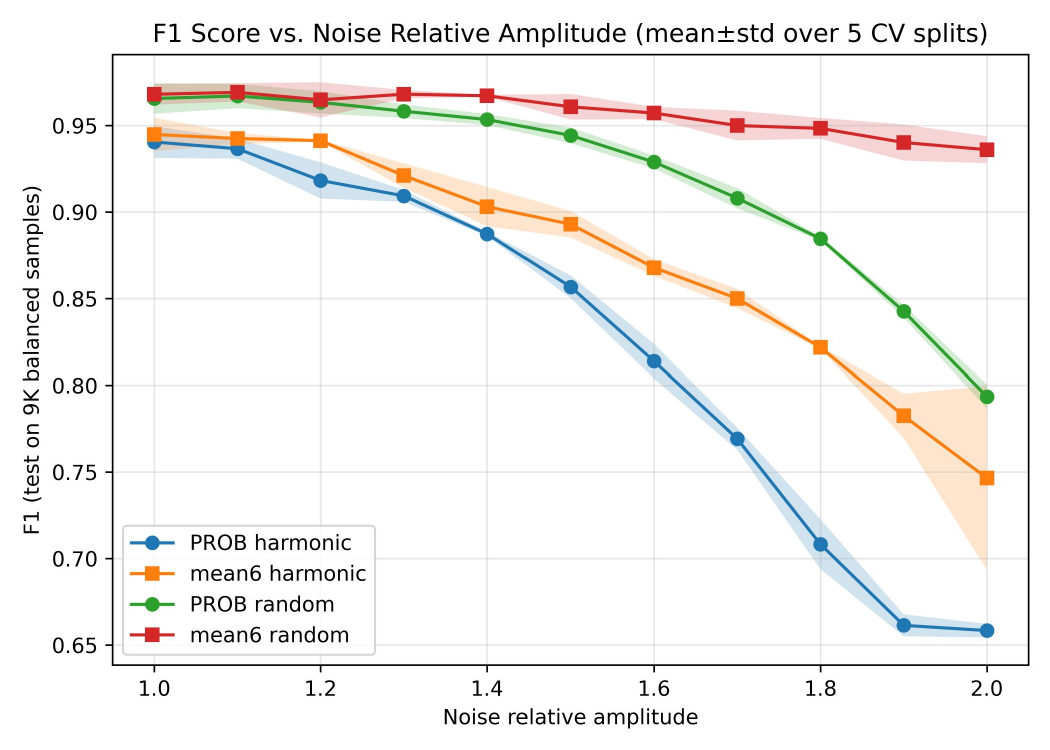}
\caption{\chdeleted{Figure 6. }F1 score in relation to the relative amplitude of noise for both harmonic and random noise injections (mean ± standard deviation across five cross-validation \chreplaced{(CV)}{} splits). The \(\tau_{\mathrm{PROB}}\) and \(\tau_{\mathrm{SHAP}}\) thresholds were adjusted individually \chreplaced{}{for criteria based only on probability and $SHAP$} using a 100-sample balanced train-set and then subsequently evaluated using a 9,000-sample balanced test set. The shaded areas around each curve show how the five CV splits differ from each other.
\chreplaced{Probability-only performance (blue, green) degrades steadily with increasing noise, while SHAP-based thresholding (orange, red; mean of 6 SHAP values) maintains higher F1 across the full range, particularly at moderate-to-high noise amplitudes, indicating a more robust decision criterion under structured and unstructured noise contamination.}
{Probability-only performance (blue, green) degrades steadily with increasing noise, while SHAP-based thresholding (orange, red; mean of 6 SHAP values) maintains substantially higher F1 across the entire range, demonstrating greater resilience to structured noise.}}
\label{fig:harm_rand}
\end{center}
\end{figure}

\chreplaced{The results are summarized in Figure~\ref{fig:harm_rand}. For each noise amplitude and each of the five cross-validation splits, we optimized the decision threshold separately for the probability-only and SHAP-based criteria on the 100-sample balanced training subset by selecting the threshold that maximized F1, and then evaluated the selected threshold on the independent 9,000-sample balanced test set. This procedure ensures that each method is assessed at its own best operating point under the same train/test protocol.
Under harmonic noise, the probability-only baseline shows a steady decline in F1 as noise amplitude increases, falling below 0.8 by relative amplitude 1.7 and approaching about 0.66 at amplitude 2.0. In contrast, the SHAP-based score remains above 0.8 through relative amplitude 1.8 and still attains about 0.75 at amplitude 2.0, corresponding to an absolute F1 advantage of roughly 0.09 at that noise level. For random noise, the SHAP-based method maintains F1 near 0.95 or higher up to approximately amplitude 1.8, whereas the probability-only rule begins to deteriorate earlier and drops to about 0.79 by amplitude 2.0.
Mechanistically, the improved robustness of SHAP-based thresholding arises because it relies on explanation-derived evidence that reflects coherent, physically meaningful attribution patterns across the three components and the P/S outputs, rather than on the raw prediction score alone. As noise increases, probability-only thresholding can still be triggered by isolated transients or unstable score fluctuations, whereas the SHAP criterion is more likely to retain detections supported by phase-consistent multi-component evidence and reject detections that lack such structure.}
{The results are summarized in Figure~6. When there is harmonic noise, the probability-only baseline shows a steady drop in F1 as the noise amplitude goes up. By relative amplitude 1.7, performance is below 0.8. The $SHAP$ criterion ("mean6 ONLY") consistently outperforms probability-only, keeping F1 above 0.8 even when noise levels are high. For random noise (green vs. red curves), the SHAP-based method again proves more resilient, sustaining near-constant F1 > 0.95 up to amplitude 1.6, whereas the probability-only rule degrades earlier.}

\chadded{To better understand the F1 trends, Figure~\ref{fig:pr_re_vs_noise} decomposes performance into precision and recall across the same range of relative noise amplitudes. The SHAP-based criterion generally maintains a more favorable precision–recall balance than probability-only thresholding, especially at moderate and high noise levels. Under harmonic noise, the probability-only baseline experiences a marked precision decline beyond relative amplitudes of about 1.6, whereas the SHAP-based rule degrades more gradually. Under random noise, SHAP-based inference remains comparatively stable in both precision and recall over a wider noise range, which is consistent with its superior F1 values in Figure~\ref{fig:harm_rand}.
An apparent increase in recall for the probability-only baseline at relative amplitudes 1.9-2.0 should not be interpreted as a genuine recovery in model robustness. Rather, it reflects a shift in the F1-optimal operating point caused by a sharp drop in the selected \(\tau_{\mathrm{PROB}}\). For example, the optimal \(\tau_{\mathrm{PROB}}\) decreases from about 0.52-0.65 at relative amplitude 1.8 to values as low as 0.09-0.50 at 1.9 and 0.13-0.28 at 2.0. This lower threshold admits many more detections, which increases recall but does so at the expense of a substantial loss in precision. In contrast, \(\tau_{\mathrm{SHAP}}\) values remain much more stable, indicating that the explanation-based criterion provides a more consistent decision rule under severe noise contamination.}

\begin{chaddedfigure}[!htbp]
\begin{center}
\includegraphics[width=0.99\textwidth]{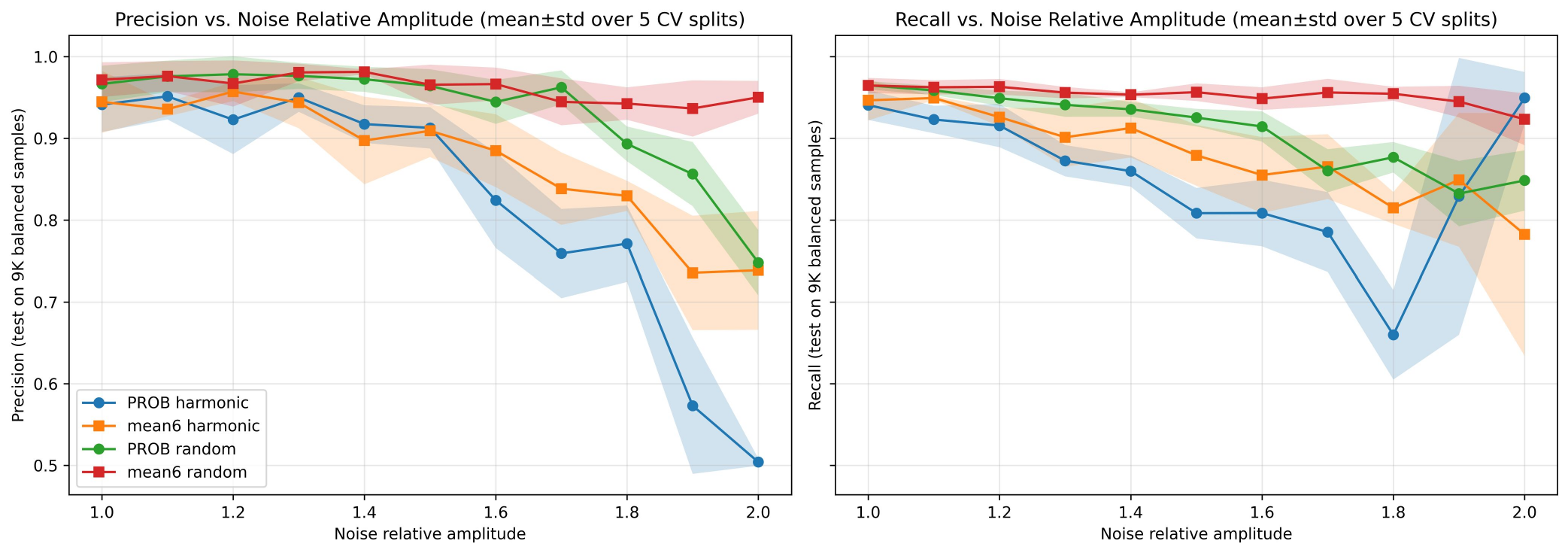}
\caption{Precision and recall as a function of relative noise amplitude for probability-only and SHAP-based thresholding (mean $\pm$ standard deviation across five cross-validation splits). Left panel shows precision; right panel shows recall. Thresholds for both criteria were optimized separately at each noise amplitude using a 100-sample balanced training set and then evaluated on a separate 9,000-sample balanced test set. Across both harmonic and random noise injections, the SHAP-based criterion generally preserves a more favorable precision–recall trade-off than probability-only thresholding as noise increases. In particular, SHAP-based inference maintains higher precision at moderate-to-high noise levels while retaining competitive recall, which explains the improved F1 behavior observed in Figure~\ref{fig:harm_rand}. The apparent recall increase for the probability-only baseline at relative amplitudes 1.9-2.0 is associated with a sharp reduction in the F1-optimal probability threshold, which shifts the operating point toward a high-recall/low-precision regime rather than indicating a genuine recovery in detector robustness.}
\label{fig:pr_re_vs_noise}
\end{center}
\end{chaddedfigure}

\begin{figure}[!htbp]
\begin{center}
\includegraphics[width=0.99\textwidth]{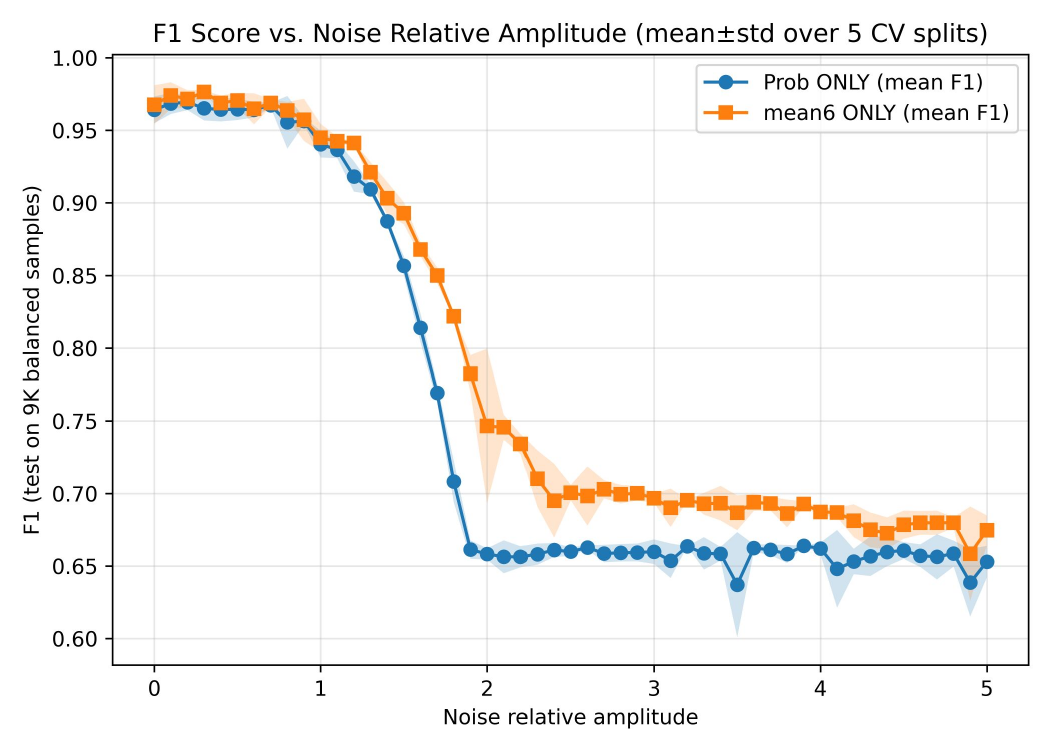}
\caption{\chdeleted{Figure 7. }F1 score versus extended noise relative amplitude (up to 5.0) for probability-only (blue) and SHAP-only (orange) thresholding, averaged over five cross-validation splits. The shaded regions around each curve indicate variability across the five CV splits. Both methods perform worse as noise levels rise, but SHAP-based inference always does better than the baseline, keeping F1 at about 0.75 at amplitude 2.0 compared to about 0.65 for probability-only. This advantage shows the practicality of using SHAP evidence as the decision rule, even when there is considerable noise.
}
\label{fig:harm_noise}
\end{center}
\end{figure}

\chreplaced{When extending the analysis to even higher noise amplitudes (up to 5×; Fig.~\ref{fig:harm_noise}), both methods inevitably lose accuracy, but SHAP-based thresholding retains a clear advantage in terms of overall F1.}
{When extending to even higher noise amplitudes (up to 5×, Fig.~7), both methods inevitably lose accuracy, but SHAP-based thresholding retains a clear advantage.}
At relative amplitudes around 2.0, the probability-only baseline collapses to F1 $\approx 0.66$, while the SHAP-based approach still holds around F1 $\approx 0.75$. This margin is useful in practice because it means that the error rate goes down by 10 to 15\% even when in highly challenging conditions.

\chreplaced{Collectively, these experiments show that SHAP-gated inference improves detection performance not only on the clean test set but also under progressively stronger noise contamination. On the clean test set, the SHAP-gated model improved the F1-score from 0.97 to 0.98 and increased recall from 0.96 to 0.97 while maintaining precision at 0.99. Across increasing harmonic and random noise levels, the SHAP-based criterion also preserved a more favorable precision-recall trade-off than probability-only thresholding, which explains its consistently higher F1 values over a broad range of relative amplitudes.
This improvement is primarily due to the fact that SHAP gating favors detections supported by coherent, physically meaningful attribution patterns across the waveform components and phase channels, rather than by a single high-probability transient. True events tend to produce stronger and more distributed SHAP evidence that is consistent with expected phase behavior, whereas noise-driven false alarms more often yield weaker or less phase-consistent attributions. At the highest noise levels, the apparent recall increase of the probability-only baseline reflects a shift in the F1-optimal operating point caused by a sharp drop in the selected \(\tau_{\mathrm{PROB}}\), which increases recall at the expense of precision rather than indicating a genuine recovery in robustness. In contrast, \(\tau_{\mathrm{SHAP}}\) remain comparatively more stable, indicating that explanation-based gating provides a more consistent decision rule under severe noise contamination.}
{Collectively, these experiments illustrate that SHAP-gated inference proves advantageous not only on "clean" test sets but also significantly improves resilience against increased noise levels. By necessitating adequate SHAP evidence, the system effectively eliminates spurious detections that would otherwise be accepted based solely on probability, while also ensuring that real events, which might be overlooked under conditions of degraded SNR, are accurately identified.}

\section{Discussion}
\label{discussion}

Our findings demonstrate that explainable AI tools can play a dual role in seismic event detection: interpreting model behavior and enhancing model performance. We showed this in the context of PhaseNet, but the approach is general and opens several avenues for further exploration.

To our knowledge, this is the first study to integrate SHAP values into the inference process of a seismic detection model. Previous works have primarily used XAI for post-hoc interpretation -- for example, visualizing that a CNN focuses on P-wave arrivals \citep{bi2021explainable, trani2022deepquake} or using LRP to debug misclassifications \citep{majstorovic2023interpreting, jiang2024explainable}. We extend those ideas by feeding the explanation back into decision-making. This places our work within the emerging field of 'explainable-by-design' performance enhancements, where the XAI component is an active part of the system pipeline, not merely a post-hoc analysis tool. 

It is possible to draw a parallel with the research conducted by ~\cite{antariksa2025xai}, who employed a SHAP-based methodology to facilitate data contamination in the training of a seismic denoising network, similarly utilizing explanations to enhance model performance prescriptively. This approach is gaining traction in related geosciences applications. For instance, \cite{Wang2025xai} applied SHAP to a 1D-CNN model for structural damage identification. Their interpretability analysis allowed them to perform optimal feature selection, creating a refined model with fewer input features that achieved significantly higher accuracy than the original. While also using SHAP, \cite{Sun2023xai} focused on diagnostic validation rather than prescriptive enhancement. They confirmed that their machine learning model for predicting Peak Ground Acceleration (PGA) learned relationships consistent with established physical laws, thereby using explainability to build trust and verify the model's scientific rationality. Our work aligns more closely with the former, using the explanation as a mechanism for direct performance improvement.

Our SHAP-gating strategy is similar to how a human analyst would check a detection: there should be a trigger (a spike in probability), and the waveform's context and features should also make sense, for example, the right amount of energy on relevant components and the appropriate length. SHAP is, in a sense, quantifying that context for the model. Our approach's success highlights the argument presented by \cite{park2024making} that a model's raw output score does not consistently serve as a dependable measure of detection confidence. 
\chadded{By demonstrating that SHAP dispersion directly correlates with SNR regimes, we provide a quantifiable metric for attribution confidence that functions independently of the raw probability score.}
Their approach involved retraining the model using specific techniques aimed at enhancing the consistency of its scores in relation to signal quality. In contrast, our approach maintains the integrity of the model while incorporating an external consistency check through SHAP. A promising avenue for future research involves integrating these methodologies: it would be beneficial to train a PhaseNet-like model incorporating a regularization term or a multi-task objective that specifically aims to enhance the SHAP mean for true events while reducing it for noise. This might directly reinforce the concept of “explanatory evidence” in the model’s learning process.

While we focused on PhaseNet (a specific CNN architecture for picking), the methodology applies to other seismic event detection models. For example, the Earthquake Transformer model, which employs self-attention and was specifically developed for regional earthquake detection, naturally generates attention weights that are subject to interpretation~\citep{mousavi2020earthquake}. It is possible to envision utilizing those attention scores in a manner akin to our \(\tau_{\mathrm{SHAP}}\) metric for the purpose of filtering outputs. For instance, it is necessary to ensure that detection focuses its attention on an arrival. Likewise, simpler CNN or LSTM-based detectors in volcano seismology~\citep{beker2022explainability} or acoustic emission monitoring could benefit from Grad-CAM or SHAP analyses to ensure they respond to physically meaningful features. Additionally, these post-hoc methods may serve as a significant benchmark for validating and comprehending the behavior of intrinsically explainable models, such as the prototype-based neural networks suggested for seismic facies classification~\citep{noh2023explainable}, thereby facilitating a comparison of various families of XAI approaches. For instance, in geophysics, Fourier Neural Operator (FNO) models \citep{li2020fourier} can be used to find anomalies in continuous seismic wavefields. FNOs are highly complex, but applying XAI to them (e.g., integrated gradients or SHAP on input frequency components) could reveal whether they’re picking up real seismic signals or artifacts. We anticipate that as more geophysical AI models come online, incorporating explainability will become a best practice to validate models before deployment.

\chadded{Although the present study reformulates PhaseNet as a binary event-versus-noise detector, the same explanation-guided principle could in principle be extended to more general seismic picking tasks. In a P- and S-wave picking setting, SHAP- or attention-based evidence would not need to act only as a binary gate; it could also be used to assess pick reliability, reject unstable or physically inconsistent picks, or flag low-confidence arrivals for analyst review. For example, a robust picking-oriented explanation metric could favor attributions that are temporally concentrated near the predicted onset and distributed across components in a way consistent with seismic phase physics, such as stronger vertical support for P arrivals and stronger horizontal support for S arrivals. In this sense, XAI could contribute not only to post-hoc interpretation, but also to confidence calibration and quality control in automatic arrival-time picking pipelines. A rigorous evaluation of this idea would require a dedicated picking study using timing-error metrics, which is beyond the scope of the present work but represents an important direction for future research.}

\chreplaced{We acknowledge several limitations of our study. First, our SHAP-gating rule was manually tuned and intentionally simple. It worked well for our balanced dataset, but in a real setting the best thresholds may shift with noise conditions, event magnitude, or waveform complexity. Recalibration or adaptive thresholding may therefore be necessary in a production environment. Future work could also investigate more advanced explanation-based metrics. For example, instead of using only the mean SHAP value across the six component-phase attributions, one could design a gating metric that combines complementary information from intermediate and deep network representations. Features from shallower layers may better capture localized onset characteristics, such as abrupt amplitude changes, short-duration transients, and fine-scale P- or S-arrival structure, whereas deeper layers may better represent broader waveform morphology, phase coherence across components, and the overall event-versus-noise pattern. A multi-layer explanation metric could therefore weight detections not only by the strength of the attribution, but also by whether the attribution is concentrated in physically meaningful arrival regions and distributed across components in a manner consistent with seismic phase behavior~\citep{li2023multilayer}. Such a strategy would improve interpretability because the gating decision would be tied more directly to identifiable waveform characteristics, and it could improve reliability by reducing acceptance of detections supported only by shallow, noise-sensitive activations or by diffuse, weak evidence.}
{We acknowledge several limitations of our study. First, our SHAP-gating rule was manually tuned and quite simple. It worked well for our balanced dataset, but in a real setting, the best thresholds might change if the noise levels change or if the characteristics of the events change (for example, bigger events might have different SHAP distributions than smaller ones). Recalibration or adaptive thresholding may be necessary in a production environment. Future research may investigate more advanced explanation-based metrics. For example, instead of just using the mean, one could come up with a metric based on the multi-layer fusion methods suggested for Grad-CAM, which combine information from different levels of the network to make a stronger explanation. This could lead to a gating criterion that is responsive to both detailed onset characteristics (from shallower layers) and overall waveform morphology (from deeper layers).}

Second, our evaluation was performed on a controlled, balanced dataset (equal numbers of signal and noise windows). In a continuous monitoring stream, the strong class imbalance (many more noise than signal windows) means that even a small increase in false positives could accumulate into frequent false triggers. In our baseline case, the SHAP‐gated rule slightly increased false positives (from 45 to 50) but significantly reduced false negatives (from 186 to 140), indicating a more sensitive detector that misses fewer true events. This improvement in recall is particularly valuable for microseismic monitoring, where the cost of missed detections often outweighs occasional extra false alarms. The modest rise in false positives remains acceptable given that SHAP evidence integrates physically meaningful features (multi-component phase energy), which should remain robust under more complex noise conditions.

\chadded{Ultimately, the combination of our statistical stability analysis and noise injection experiments contextualizes the practical value of explanation-based decision rules. While the absolute F1 improvement achieved by the SHAP-gated inference on our high-quality clean dataset is relatively modest (+0.01), this simply reflects a baseline model already operating near its performance ceiling, where the SHAP rule effectively filters out the remaining marginal false negatives. The true operational value of this approach lies in its behavior under real-world signal degradation. As demonstrated, when raw model probabilities collapse under heavy noise contamination~(e.g., dropping to $F1 \approx 0.66$ at a relative amplitude of 2.0), the multi-component physical consistency required by the SHAP gating prevents catastrophic failure, maintaining an $F1 \approx 0.75$. This confirms that XAI-driven decision rules function less as a tool for marginal gains in pristine data, and more as a critical safety net for autonomous monitoring in challenging deployment environments.}

\section{Conclusions}
\label{conclusions}

\chreplaced{This study shows that explainable AI can be used for both model interpretation and improving detection in microseismic monitoring. We applied Grad-CAM to the PhaseNet model and found that the network’s strongest activations generally align with the P- and S-wave arrival regions, indicating that its decisions are broadly consistent with geophysical expectations. This agreement was assessed qualitatively in the present study; a more formal validation based on overlap metrics between attribution regions and manually annotated arrival windows would be a useful direction for future work. We also note that the correspondence is not equally sharp in all cases: while high-SNR events show concentrated activation near the arrivals, low-SNR cases can exhibit broader and less localized attribution patterns, and noise windows may show weak scattered responses. Along with this, SHAP analysis provided quantitative, component-level attributions that were consistent with established seismic wave propagation physics, strengthening confidence in the model’s internal logic.}
{This study shows that explainable AI can be used for both model interpretation and improving detection in microseismic monitoring. We applied Grad-CAM to the PhaseNet model and visually verified that the network's decision-making process is geophysically valid, concentrating on the P- and S-wave arrivals, confirming that a seismic event happened. Along with this, SHAP analysis gave quantitative, component-level attributions that are consistent with established physics of seismic wave propagation, which makes the model's internal logic even more reliable.}

\chreplaced{More importantly, we have shown that these explanations can be actively incorporated into the inference pipeline to make the detector more robust. On the clean test set, the SHAP-gated inference scheme improved the F1-score from 0.97 to 0.98 and reduced the number of false negatives from 186 to 140 while maintaining precision at 0.99. This improvement arises because the SHAP-based criterion favors detections supported by coherent, physically meaningful attribution patterns across waveform components and phase channels, helping preserve genuine low-SNR events that may receive lower raw probability scores while filtering detections that are not supported by stable phase-consistent evidence. More broadly, although demonstrated here with PhaseNet, the same explanation-guided inference principle could be adapted to other geophysical and time-series models, including CNN-, LSTM-, or Transformer-based detectors, provided that a reliable attribution or attention-based evidence measure can be defined. 
Beyond binary event detection, the same explanation-guided strategy may also prove useful for automatic P- and S-wave picking by providing an additional measure of pick reliability and physical consistency.
This illustrates a significant new pathway for XAI in the geosciences, shifting from mere interpretation to proactive performance improvement.}
{More importantly, we have shown that these explanations can be actively added to the inference pipeline to make the detector more robust. Our new SHAP-gated inference scheme, which combines the model's output probability with a measure of explanatory evidence, improved the F1-score by lowering the number of false negatives. This illustrates a significant new pathway for XAI in the geosciences, shifting from mere interpretation to proactive performance improvement.}

The enhanced resilience to noise demonstrated by the SHAP-gated model indicates that this method holds significant promise for practical monitoring situations where signal quality may fluctuate considerably. The principles of explanation-guided inference, while illustrated using PhaseNet, are widely applicable to various deep learning models in seismology and other fields. Ultimately, by making AI models more transparent and leveraging their explanations to make them more reliable, we can accelerate the confident deployment of AI in critical geoscience applications.

\section*{Declaration of competing interest}

The authors declare that they have no known competing financial interests or personal relationships that could have appeared to influence the work reported in this paper.

\section*{Data availability}

The microseismic dataset underlying this article was provided by Seismik s.r.o. to be used in this study. This dataset will be shared on request to the corresponding author with permission of Seismik s.r.o.

\section*{Acknowledgments}

The authors would like to acknowledge the support provided by the Deanship of Research (DR) at King Fahd University of Petroleum \& Minerals (KFUPM) for funding this work through project No. MbSC2601.


\section*{Declaration of generative AI and AI-assisted technologies in the manuscript preparation process}

During the preparation of this work, the authors used ChatGPT (OpenAI, USA) to refine the language and improve the readability of the manuscript. After using this tool, the authors thoroughly reviewed and edited the content as needed and take full responsibility for the content of the published article.

\bibliographystyle{elsarticle-harv}
\bibliography{ref_aig}

@article{myren2025evaluation,
    author = {Myren, Samuel and Parikh, Nidhi and Rael, Rosalyn and Flynn, Garrison and Higdon, Dave and Casleton, Emily},
    title = {Evaluation of Seismic Artificial Intelligence with Uncertainty},
    journal = {Seismological Research Letters},
    year = {2025},
    month = {07},
    issn = {0895-0695},
    doi = {10.1785/0220240444},
}

@inproceedings{beker2022explainability,
  title={Explainability analysis of CNN in detection of volcanic deformation signal},
  author={Beker, Teo and Ansari, Homa and Montazeri, Sina and Song, Qian and Zhu, Xiao Xiang},
  booktitle={IGARSS 2022-2022 IEEE International Geoscience and Remote Sensing Symposium},
  pages={4851--4854},
  year={2022},
  organization={IEEE}
}

@Article{Sun2023xai,
AUTHOR = {Sun, Rui and Qi, Wanwan and Zheng, Tong and Qi, Jinlei},
TITLE = {Explainable Machine-Learning Predictions for Peak Ground Acceleration},
JOURNAL = {Applied Sciences},
VOLUME = {13},
YEAR = {2023},
NUMBER = {7},
ARTICLE-NUMBER = {4530},
URL = {https://www.mdpi.com/2076-3417/13/7/4530},
ISSN = {2076-3417},
DOI = {10.3390/app13074530}
}

@article{Wang2025xai,
author = {Wang, Xinwei and Wei, Zheng and Wang, Zhihao and Wei, Shuaiqiang and Li, Yanchun and Shahzad, Muhammad Moman},
title = {Explainable AI-Driven Optimal Feature Selection for the Identification of Structural Damage},
journal = {Structural Control and Health Monitoring},
volume = {2025},
number = {1},
pages = {7253150},
doi = {https://doi.org/10.1155/stc/7253150},
url = {https://onlinelibrary.wiley.com/doi/abs/10.1155/stc/7253150},
eprint = {https://onlinelibrary.wiley.com/doi/pdf/10.1155/stc/7253150},
year = {2025}
}

@article{li2023multilayer,
  title={Multilayer Grad-CAM: An effective tool towards explainable deep neural networks for intelligent fault diagnosis},
  author={Li, Sinan and Li, Tianfu and Sun, Chuang and Yan, Ruqiang and Chen, Xuefeng},
  journal={Journal of manufacturing systems},
  volume={69},
  pages={20--30},
  year={2023},
  publisher={Elsevier}
}

@article{yoo2022vibration,
  title={Vibration analysis process based on spectrogram using gradient class activation map with selection process of CNN model and feature layer},
  author={Yoo, Youngjun and Jeong, Seongcheol},
  journal={Displays},
  volume={73},
  pages={102233},
  year={2022},
  publisher={Elsevier}
}

@article{antariksa2025xai,
  title={XAI-driven contamination for self-supervised denoising with pixel-level anomaly detection in seismic data},
  author={Antariksa, Gian and Koeshidayatullah, Ardiansyah and Das, Subasish and Lee, Jihwan},
  journal={Journal of Applied Geophysics},
  volume={238},
  pages={105723},
  year={2025},
  publisher={Elsevier}
}

@article{jena2023explainable,
  title={Explainable artificial intelligence (XAI) model for earthquake spatial probability assessment in Arabian peninsula},
  author={Jena, Ratiranjan and Shanableh, Abdallah and Al-Ruzouq, Rami and Pradhan, Biswajeet and Gibril, Mohamed Barakat A and Khalil, Mohamad Ali and Ghorbanzadeh, Omid and Ganapathy, Ganapathy Pattukandan and Ghamisi, Pedram},
  journal={Remote Sensing},
  volume={15},
  number={9},
  pages={2248},
  year={2023},
  publisher={MDPI}
}

@article{guo2023interpretable,
  title={An interpretable deep learning method for identifying extreme events under faulty data interference},
  author={Guo, Jiaxing and Tang, Zhiyi and Zhang, Changxing and Xu, Wei and Wu, Yonghong},
  journal={Applied Sciences},
  volume={13},
  number={9},
  pages={5659},
  year={2023},
  publisher={MDPI}
}

@inproceedings{edigbue2025explaining,
  title={Explaining Deep Learning Models in Full Waveform Inversion: Enhancing Transparency in Seismic Data Interpretation},
  author={Edigbue, Paul and Al-Shuhail, Abdullatif and Hanafy, Sherif},
  booktitle={SPE Middle East Oil and Gas Show and Conference},
  pages={D021S047R005},
  year={2025},
  organization={SPE}
}

@article{noh2023explainable,
  title={Explainable deep learning for supervised seismic facies classification using intrinsic method},
  author={Noh, Kyubo and Kim, Dowan and Byun, Joongmoo},
  journal={IEEE Transactions on Geoscience and Remote Sensing},
  volume={61},
  pages={1--11},
  year={2023},
  publisher={IEEE}
}

@article{lundberg2017unified,
  title={A unified approach to interpreting model predictions},
  author={Lundberg, Scott M and Lee, Su-In},
  journal={Advances in neural information processing systems},
  volume={30},
  year={2017}
}

@inproceedings{selvaraju2017grad,
  title={Grad-cam: Visual explanations from deep networks via gradient-based localization},
  author={Selvaraju, Ramprasaath R and Cogswell, Michael and Das, Abhishek and Vedantam, Ramakrishna and Parikh, Devi and Batra, Dhruv},
  booktitle={Proceedings of the IEEE international conference on computer vision},
  pages={618--626},
  year={2017}
}

@inproceedings{jiang2024explainable,
  title={Explainable AI for transparent seismic signal classification},
  author={Jiang, Jiaxin and Stankovic, Vladimir and Stankovic, Lina and Murray, David and Pytharouli, Stella},
  booktitle={IGARSS 2024-2024 IEEE International Geoscience and Remote Sensing Symposium},
  pages={8801--8805},
  year={2024},
  organization={IEEE}
}

@article{majstorovic2023interpreting,
  title={Interpreting convolutional neural network decision for earthquake detection with feature map visualization, backward optimization and layer-wise relevance propagation methods},
  author={Majstorovi{\'c}, Josipa and Giffard-Roisin, Sophie and Poli, Piero},
  journal={Geophysical Journal International},
  volume={232},
  number={2},
  pages={923--939},
  year={2023},
  publisher={Oxford University Press}
}

@article{trani2022deepquake,
  title={DeepQuake—An application of CNN for seismo-acoustic event classification in The Netherlands},
  author={Trani, Luca and Pagani, Giuliano Andrea and Zanetti, Jo{\~a}o Paulo Pereira and Chapeland, Camille and Evers, L{\"a}slo},
  journal={Computers \& Geosciences},
  volume={159},
  pages={104980},
  year={2022},
  publisher={Elsevier}
}

@article{bi2021explainable,
  title={Explainable time--frequency convolutional neural network for microseismic waveform classification},
  author={Bi, Xin and Zhang, Chao and He, Yao and Zhao, Xiangguo and Sun, Yongjiao and Ma, Yuliang},
  journal={Information Sciences},
  volume={546},
  pages={883--896},
  year={2021},
  publisher={Elsevier}
}

@inproceedings{bedle2024application,
  title={Application of vector plots, LIME, and SHAP for seismic facies machine learning evaluation},
  author={Bedle, Heather and Lubo-Robles, David},
  booktitle={SEG International Exposition and Annual Meeting},
  pages={SEG--2024},
  year={2024},
  organization={SEG}
}

@article{lubo2022quantifying,
  title={Quantifying the sensitivity of seismic facies classification to seismic attribute selection: An explainable machine-learning study},
  author={Lubo-Robles, David and Devegowda, Deepak and Jayaram, Vikram and Bedle, Heather and Marfurt, Kurt J and Pranter, Matthew J},
  journal={Interpretation},
  volume={10},
  number={3},
  pages={SE41--SE69},
  year={2022},
  publisher={Society of Exploration Geophysicists and American Association of Petroleum~…}
}

@inproceedings{saikia2019seismic,
  title={Seismic signal interpretation for reservoir facies classification},
  author={Saikia, Pallabi and Nankani, Deepankar and Baruah, Rashmi Dutta},
  booktitle={International Conference on Pattern Recognition and Machine Intelligence},
  pages={409--417},
  year={2019},
  organization={Springer}
}

@article{park2024making,
  title={Making Phase-Picking Neural Networks More Consistent and Interpretable},
  author={Park, Yongsoo and Delbridge, Brent G and Shelly, David R},
  journal={The Seismic Record},
  volume={4},
  number={1},
  pages={72--80},
  year={2024},
  publisher={Seismological Society of America}
}

@article{mousavi2020earthquake,
  title={Earthquake transformer—an attentive deep-learning model for simultaneous earthquake detection and phase picking},
  author={Mousavi, S Mostafa and Ellsworth, William L and Zhu, Weiqiang and Chuang, Lindsay Y and Beroza, Gregory C},
  journal={Nature communications},
  volume={11},
  number={1},
  pages={3952},
  year={2020},
  publisher={Nature Publishing Group UK London}
}

@article{zhu2019phasenet,
  title={PhaseNet: a deep-neural-network-based seismic arrival-time picking method},
  author={Zhu, Weiqiang and Beroza, Gregory C},
  journal={Geophysical Journal International},
  volume={216},
  number={1},
  pages={261--273},
  year={2019},
  publisher={Oxford University Press}
}

@article{li2020fourier,
  title={Fourier neural operator for parametric partial differential equations},
  author={Li, Zongyi and Kovachki, Nikola and Azizzadenesheli, Kamyar and Liu, Burigede and Bhattacharya, Kaushik and Stuart, Andrew and Anandkumar, Anima},
  journal={arXiv preprint arXiv:2010.08895},
  year={2020}
}

\iftoggle{chfinal}{}{%
\newpage
\clearpage
\listofchanges}

\end{document}